%% file: Main.tex
\pdfoutput=1
\PassOptionsToPackage{table}{xcolor}
\documentclass[11pt]{article}

\usepackage{EMNLP2023}

\usepackage{times}
\usepackage{latexsym}

\usepackage{CJK}
\usepackage[table]{xcolor}
\usepackage{adjustbox}
\usepackage{bbding}
\usepackage{tabularx}
\usepackage{bm}
\usepackage{amssymb}
\usepackage{booktabs}
\usepackage{url}
\usepackage{multicol}
\usepackage{tabularx}
\usepackage{booktabs}
\usepackage{color}
\usepackage{arydshln}
\usepackage{multirow}
\usepackage{amsmath} 
\usepackage{bm}
\usepackage{subfigure}
\usepackage{amstext}
\usepackage{tikz}
\usepackage{pgfplots}
\usepackage{hyperref}
\usepackage{subfigure}
\usepackage{enumitem}
\usepackage{mathtools}

\usepackage{stfloats}

\input{macros}

\usepackage[T1]{fontenc}

\usepackage[utf8]{inputenc}

\usepackage{microtype}

\usepackage{inconsolata}

 \usepackage{graphicx}
\title{\textsc{ReNoVi}: A Benchmark Towards Remediating Norm Violations in Socio-Cultural Conversations}

\author{Haolan Zhan\textsuperscript{\rm \heart},  Zhuang Li\textsuperscript{\rm \heart}, Xiaoxi Kang\textsuperscript{\rm \heart}, Tao Feng\textsuperscript{\rm \heart}, Yuncheng Hua\textsuperscript{\rm \heart},
\textbf{Lizhen Qu}\textsuperscript{\rm \heart}, \\
\textbf{Yi Ying}\textsuperscript{\rm \club}, \textbf{Mei Rianto Chandra}\textsuperscript{\rm \club}, \textbf{Kelly Rosalin}\textsuperscript{\rm \club}, \textbf{Jureynolds Jureynolds}\textsuperscript{\rm \club}, \\
\textbf{Suraj Sharma}\textsuperscript{\rm \diamondsmall}, \textbf{Shilin Qu}\textsuperscript{\rm \heart}, \textbf{Linhao Luo}\textsuperscript{\rm \heart}, \textbf{Lay-Ki Soon}\textsuperscript{\rm \heart}, \\
\textbf{Zhaleh Semnani Azad}\textsuperscript{\rm \diamondsmall}, \textbf{Ingrid Zukerman}\textsuperscript{\rm \heart}, \textbf{Gholamreza Haffari}\textsuperscript{\rm \heart} \\
\textsuperscript{\rm \heart} Monash University, Australia \& Malaysia\\
\textsuperscript{\rm \diamondsmall}  California State University, Northridge, CA 
\textsuperscript{\rm \club} Binus University, Indonesia \\
\{firstname.lastname\}@monash.edu, \{suraj.sharma, zhaleh.semnaniazad\}@csun.edu\\ 
}

\begin{document}

\maketitle

\input{0-abstract}

\input{1_intro_lizhen}
\input{2-background}

\input{2-framework}
\input{3-dataset}
\input{4-experiments}

\input{5-relatedwork}
\input{6-conclusion}
\input{7-limitations}

\input{8-Ethics}

\section*{Acknowledgement}

This material is based on research sponsored by DARPA under agreement number HR001122C0029. The U.S. Government is authorized to reproduce and distribute reprints for Governmental purposes notwithstanding any copyright notation thereon. 

\bibliography{anthology,custom,ref}
\bibliographystyle{acl_natbib}

\clearpage

\appendix

\input{9-appendix}

\end{document}

%% file: macros.tex
\newcommand{\eg}{e.g., }

\newcommand{\ourdata}{\textsc{ReNoVi}}

\newcommand{\heart}{\ensuremath\heartsuit}
\newcommand{\diamondsmall}{\ensuremath\diamondsuit}
\newcommand{\club}{\ensuremath\clubsuit}

\definecolor{dgreen}{rgb}{0,0.55,0}
\definecolor{mgreen}{rgb}{0,0.7,0}


%% file: 0-abstract.tex
\begin{abstract}



Norm violations occur when individuals fail to conform to culturally accepted behaviors, which may lead to potential conflicts. 
Remediating norm violations requires social awareness and cultural sensitivity of the nuances at play.
To equip interactive AI systems with a remediation ability,  we offer \ourdata\ --- a large-scale corpus of 9,258 multi-turn dialogues annotated
with social norms, as well as define a sequence of tasks to help understand and remediate norm violations step by step.
\ourdata\ consists of two parts: 512 human-authored dialogues (\textit{real data}), and 8,746 
synthetic conversations generated by ChatGPT through prompt learning. 
While collecting sufficient human-authored data is costly, synthetic conversations provide suitable amounts of data to help mitigate the scarcity of training data, as well as the chance to assess the alignment between LLMs and humans in the awareness of social norms.
We thus harness the power of ChatGPT to generate synthetic training data 
for our task. To ensure the quality of both human-authored and synthetic data, {we follow a quality 
control protocol during data collection.} Our experimental results demonstrate the importance of remediating norm 
violations in socio-cultural conversations, as well as the improvement in performance obtained from synthetic data\footnote{Sampled cases and codes can be found in the appendix, and we will release the full dataset upon publication.}.

\end{abstract}


%% file: 1_intro_lizhen.tex
\section{Introduction}
Social norms, the informal rules that define acceptable and appropriate behavior in groups or societies, are extensively studied by sociologists, anthropologists and psychologists for interpersonal communication~\citep{sep-social-norms}. Expectancy Violation theory (EVT) and its extensions discuss the effects of norm or behavior violations on interpersonal communication outcomes~\citep{burgoon2005cross,burgoon2015expectancy}. According to the theory and empirical studies, 
violations of social norms often invoke punishment, such as costly sanctions, confrontation, gossip and social exclusion~\citep{molho2020direct}.

\input{figures/intro-framework}

Large language models (LLMs) demonstrate reasoning and generalization capabilities that help people with a variety of communication tasks, \eg essay writing and customer support. However, little is known about how LLMs align with human interpretations of social norms and how they can assist humans with socio-cultural verbal communication. This work aims to benchmark LLMs' ability to understand the influence of negative norm violations caused by human behaviors and mitigate their potential harm. The closest work to ours~\citep{liu2023trainingSociety} investigates the alignment between LLMs and humans in terms of general social values, such as honesty and harmlessness, without norms pertaining to a culture. Other studies
focus on extracting unknown norm rules~\cite{fung2022normsage}, recognizing their status (\textit{adherence} or \textit{violation}) and associated social factors~\citep{zhan2023socialdial}, and normative reasoning~\cite{forbes2020social}. 

To achieve our goals, we construct a novel benchmark, called \ourdata, to evaluate LLMs on assisting humans with \underline{re}mediating negative \underline{no}rm \underline{vi}olations in textual conversations. As illustrated in Fig.~\ref{fig:intro}, LLMs need to complete a sequence of four main tasks: 
(1)~detect negative norm violations, 
(2)~estimate impact of violations, 
(3)~generate remediation measures, and 
(4)~justify the generated measures and convey relevant knowledge of social norms. 
The latter two tasks are grounded in Interaction Adaptation Theory (IVT)~\citep{ebesu2015IAT}, which explains how, when and why interlocutors adjust their behavior in interpersonal communication.  We choose Chinese culture for this benchmark as China is a populous country and an important commercial partner.  

Our dataset consists of 9,258 multi-turn dialogues, including 512 human-authored conversations, 
and 8,746 synthetic conversations generated by ChatGPT\footnote{\url{https://chat.openai.com/}}. We use synthetic conversations because (i)~they help mitigate the scarcity of training data for improving the quality of open-source LLMs such as privacy-sensitive applications, and (ii)~they can be used to assess the alignment between ChatGPT and humans in terms of social norms. We conduct extensive analyses  and experiments to explore the differences between human-authored and synthetic conversations.


On \ourdata, we conduct the {first} empirical study using a variety of LLMs and offer the following findings:
\begin{itemize}
    \item We observe that solely relying on synthetic data doesn't enhance the model's performance. However, merging synthetic data with a small amount of human-authored data can enhance violation detection performance. 
    \item Quantitative and human evaluation demonstrates the potential of LLMs to align with human capabilities in awareness of social norms.
\end{itemize}


%% file: figures/intro-framework.tex
\begin{figure*}
    \centering
    \includegraphics[width=1\textwidth]{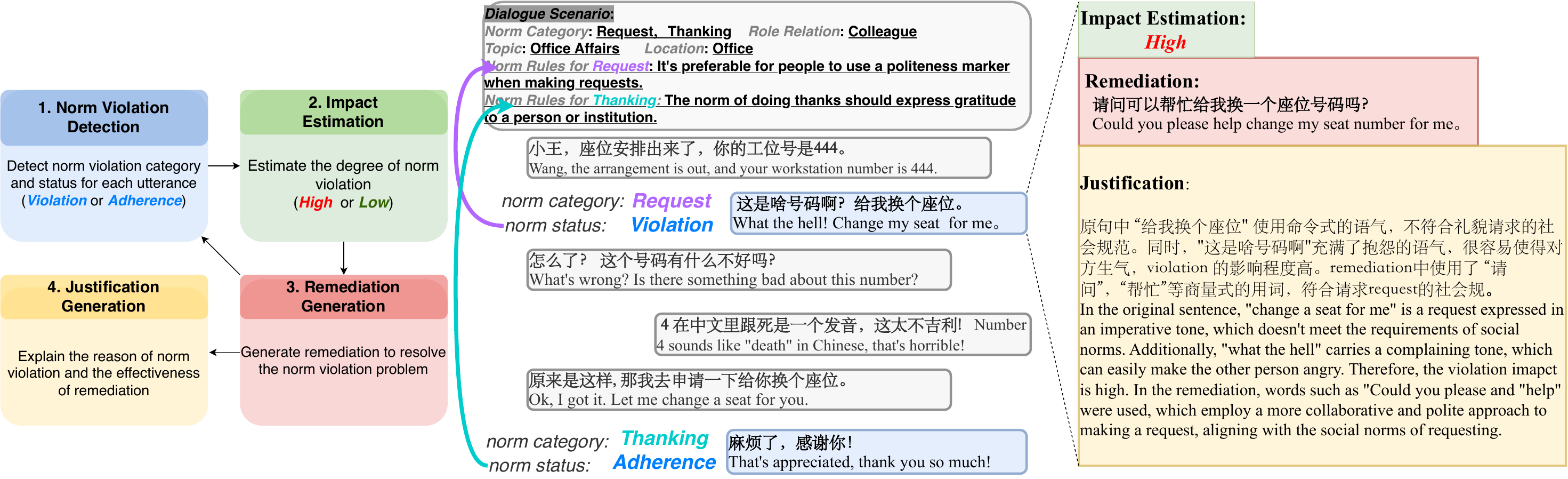}
    \caption{Main tasks of our framework (left): (1)~norm violation detection, (2)~violation impact estimation, (3)~remediation generation, and (4)~justification generation. Each dialogue (right) contains a corresponding dialogue scenario related to social norms. The detailed norm categories and rules are presented in Appendix~\ref{sec:def-social-norm}.}
    \label{fig:intro}
\end{figure*}

%% file: 2-background.tex
\section{Background}
In this section, we provide a brief introduction of EVT and IAT, as they lay the foundation of understanding human behaviors in terms of social norms during interpersonal communication.

\paragraph{Expectancy Violations Theory.}
EVT is a useful theory in the social sciences that can inform 
\textit{\textbf{how norm violations are detected}} and evaluated~\citep{burgoon1993interpersonal,burgoon2005cross}. Expectancies are enduring normative patterns of behaviors that are anticipated during interactions~\citep{burgoon1990nonverbal}. Different cultures evolve different expectancies due to their unique histories and priorities~\cite{chiu2010intersubjective}. When a behavior is perceived to be sufficiently discrepant from what was expected, an expectancy violation occurs~\cite{burgoon1993interpersonal}. The interpretations and evaluations of violations determine whether they are positive or negative, and a negative violation usually causes damage.
The \textit{\textbf{effect of a violation}} is determined based on how it was appraised. Violations are appraised with a valence and intensity, depending on many variables such as who committed the violation, where it occurred, and how important the violated norm is. We formulate the analysis on the effects of violations by categorizing them into high or low impact.

\paragraph{Interaction Adaptation Theory.}
IAT is a theory that extends EVT to be more comprehensive in accounting for concurrent interactions by emphasizing the entrainment between interlocutors during normal interactions~\cite{burgoon1997researching}. We use this theory to better understand \textit{\textbf{how remediation occurs and is facilitated following a norm violation}}. One of the principles of IAT is that during conversations, a pressure for matching and reciprocity exists~\cite{burgoon2005cross}. In other words, people exhibit highly similar nonverbal and verbal communication patterns when interacting. These behaviors are important given the necessity for people to signal common ground during interactions. Matching refers to similarities in linguistic and nonverbal behaviors, while reciprocity refers to the changes individuals exhibit during interactions to achieve greater similarity with their interaction partners.
We apply this principle to the remediation task. 






%% file: 2-framework.tex
\section{Task Definitions}
\label{sec:task_def}
We operationalize EVT and IAT for analyzing and mitigating negative norm violations into the following tasks ordered by their dependencies. 

\paragraph{Task~1: Norm Violation Detection.} Given an utterance associated with social norms of interest, the task determines (1) which \textit{\textbf{norm category}} it belongs to and (2) whether it \textit{\textbf{adheres to}} or \textit{\textbf{violates}} the corresponding norm rules\footnote{Herein, we consider only negative violations in Chinese culture. Positive violations are not observed in our collected conversations. } (as shown in Figure~\ref{fig:intro}). 
We specialize social norms in 7 typical scenarios in daily life. Following~\citep{zhan2023socialdial}, we categorize norm scenarios by people's intents and include all categories in their work, as well as two novel categories: \textit{thanking} and \textit{leave-taking}. Details about norm categories and corresponding norm rules can be found in Appendix~\ref{sec:def-social-norm}.

\paragraph{Task~2: Impact Estimation.} One important aspect of EVT is to predict interaction outcomes of violations, whether interactions should be involving, unpleasant, disinterest etc.. After experimenting with different annotation schemas, we opt to divide the effect of a violation into \textit{\textbf{high impact}} and \textit{\textbf{low impact}}, in order to achieve high agreement among annotators. The impact of a violation is considered as high if it likely leads to serious consequences, such as disengagement, negative emotions of the interlocutor or even damage to the relationship between interlocutors, otherwise its impact is low. 

\paragraph{Task~3: Remediation Generation.} According to IAT, behavior matching and reciprocity is expected following a perceived violation. Therefore, for LLMs, remediation measures can be generated to either rephrase the norm-violating ones in order to change their status to adherence, or provide instructions regarding how to conform to the corresponding norms. Figure~\ref{fig:intro} shows an example of the former case by rephrasing the utterance more politely. The latter case is useful when e.g. a decision needs to be changed from invitation rejection to invitation acceptance. This task can be used to study the alignment between LLMs and humans following violations. 

\paragraph{Task~4: Justification Generation.} This task is suggested largely from a practical perspective, because it may avoid recurred violations by teaching users the relevant norm rules and explaining why the remediation measures are effective. From a technical perspective, the task encourages models to be more explainable and provides a way to verify to what degree generated remediations align with human behaviors as well as the theories, e.g. IAT.


%% file: 3-dataset.tex
\section{\ourdata\ Dataset}



We introduce the \ourdata\ dataset, which contains 9,258 multi-turn dialogue instances with fine-grained annotation labels. 
To the best of our knowledge, \ourdata\ is the first dataset used to explore the remediation of norm violations based 
on Chinese cultural norms. 
In the rest of this section, we explain data collection (\S4.1), data quality control (\S4.2), data summary statistics (\S4.3), and comparisons of human-authored v.s. synthetically generated data (\S4.4).

\subsection{Data Collection}
We explain collecting human-authored (\S4.1.1) and synthetically-generated (\S4.1.2) dialogue data. 
\subsubsection{Curation of Human-authored Dialogues}

\noindent \textbf{Annotator Training and Examination.}
\label{sec:training}
Dialogue instances in our dataset are highly related to Chinese social norms. We,  therefore, 
invited 20 university lecturers and students who are familiar with Chinese culture to the annotation training procedure. 
To ensure that these crowd-workers provide effective social-cultural dialogues annotated with appropriate 
remediation and justifications, we designed a training tutorial. In the training tutorial, we decomposed the 
crowd-sourcing process into two stages: \textit{dialogue curation} and \textit{post annotation}. After participants
finish the tutorial, they are required to take an exam, and they proceed to the dialogue curation stage only 
if they pass the exam. At the end of this process, we had 15 crowd-workers.

\noindent \textbf{Preparation of Dialogue Scenarios.}
To encourage the crowd-workers to incorporate relevant social norms, we prepared an initial dialogue scenario for each potential dialogue. Each dialogue scenario contains a set of attributes: 1) \textit{location}, 2) \textit{role relationship}, 3) \textit{topic} and 4) \textit{social norms} including norm category and norm rules. For example, as shown in Figure~\ref{fig:intro}, the dialogue scenario show that the two interlocutors (with a specific \textit{role relation}) should talk about a \textit{topic} at a \textit{location}. Relevant social norms in this dialogue include \textit{request} and \textit{thanking}, as well as their corresponding norm rules. 

\noindent \textbf{Dialogue Curation.}
In the next stage, we instructed each crowd-worker to write a dialogue for each provided initial set of social factors. When writing a dialogue, crowd-workers were required to consider the following constraints: (1) each dialogue should contain all the social factors in the initial set; (2) for each dialogue, there
should be at least one utterance that violates or adheres to the norm rule, e.g., as seen in Figure~\ref{fig:intro}, the second utterance \textit{\textcolor{blue}{violates}} the social norm \textit{request} and the last utterance \textit{\textcolor{blue}{adheres}} to the social norm \textit{thanking}; (3) minimum 8 utterances for each dialogue.

\noindent \textbf{Post Annotation.}
In the post-annotation stage, we asked the same groups of crowd-workers to complete the annotation tasks based on their written dialogues. These annotation categories include  1) \textit{social norm} for each utterance including \textit{norm category} and \textit{violation status}. If the violation status of a utterance is \textit{True}, the following labels should be annotated: 2) \textit{impact estimation} to evaluate the effect of violations, 3) \textit{remediation} for each violation utterance, and 4) \textit{justification} for each violation utterance. The statistical distributions of each norm category can be found in Figure~\ref{fig:statistics}.

\subsubsection{Curation of Synthetic Dialogues}

The collection of human-authored data is expensive and time-consuming. Our motivation for synthetic data collection is two-fold: 1) expediting the process by acquiring ample data at a reduced expense, and 2) providing the chance to assess the alignment between ChatGPT and humans in awareness of social norms. Therefore, we leverage powerful instruction-following  LLMs (e.g., ChatGPT) to generate synthetic dialogues at scale. The curation of synthetic dialogues is formulated in two stages: 1) synthetic dialogue generation with annotations and 2) remediation and justification generation.

\noindent \textbf{Synthetic Dialogue Generation with Annotations.}
Similar to the human-authored dialogues, we prepared a scenario for each synthetic dialogue. Then, we present ChatGPT with our ontology-based prompts. The prompts incorporate the ontology labels in the scenario (e.g., location, topic) and instructions to generate synthetic dialogues. Besides, to annotate the dialogue automatically with the labels of violation status, the prompt also includes instructions to ask ChatGPT to annotate each utterance automatically. We present a detailed example in Appendix~\ref{synthetic_ex_app}.


\noindent \textbf{Remediation and Justification Generation.}
By prompting ChatGPT with the dialogue context, violating utterances and the rules of social norms, the LLM can then automatically generate synthetic remediation and justification. 
We take a zero-shot prompting approach, where we prompt ChatGPT to identify utterances that violate social norms and request it to rewrite those utterances, while considering the contextual information. The goal is to produce a revised form of norm-violating utterance that aligns with social norms. Additionally, ChatGPT is expected to provide a rationale (justification) for why the newly generated utterance does not violate our predefined social norms.

\input{figures/process-synthetic-collection}

\input{tables/overall_statistics}

\input{tables/label_statistics}

\input{figures/distribution_discrepancy}

\subsection{Quality Control}

To ensure data quality, we reviewed all 512 human-authored dialogues and 500 sampled synthetic dialogues. For both human-authored and  synthetic dialogues, we conducted careful checks for our proposed  tasks in \S3.
For the labels in norm category and violation status (task 1) and violation impact estimation (task 2), we asked two other quality inspectors (in addition to the previous annotators) to review these labels. We calculated the inter-annotator agreement~\cite{cohen1960coefficient}, and the Kappa score for  norm category, violation status, and violation impact are 0.55, 0.68, and 0.59, respectively.

We further conducted external reviews on those annotated labels where quality inspectors did not reach an agreement. Finally, among these reviewed 1,012 dialogues (512 human-authored + 500 synthetic), 28\%  (283 dialogues) did not reach an agreement, comprising 97 human-authored dialogues and 186 synthetic dialogues. All these 283 divergent dialogues were sent to the chief annotator (who created the training protocol in \S~\ref{sec:training}) to  conduct the final revision on the labels.


\subsection{Statistics Summary}

The overall statistics of \ourdata\ dataset are shown in Table~\ref{tab:dialogue_stat} for 512 human-authored dialogues and 8,746 synthetic instances. Relatively long conversations indicate that \ourdata\ provides effective multi-turn dialogues to explore norm violation and remediation issues in real scenarios (Avg. 15.29 utterances for each human-authored dialogue), which is longer than previous single context-response pair-wised datasets~\cite{forbes2020social,ziems2022moral, 10.1162/tacl_a_00561} or vanilla violation detection dataset (Avg. 6.63 utterances)~\cite{zhan2023socialdial}.
Besides, we notice that the average length of sentences (e.g, dialogue utterance, remediation and justification) generated by ChatGPT is longer than those human-authored ones. The major difference is that humans write concise and succinct sentences, while machine-generated sentences are more detailed and comprehensive.

We also present the statistics of norm category annotations in Figure~\ref{fig:statistics}. As seen, ``\textit{request}'' label has the highest proportion in both human-authored and synthetic data, indicating that it's one of the most widely used norms in dialogues. Meanwhile, the distributions of other categories are different between human-authored and synthetic data. For instance, synthetic data augments the ``\textit{criticism}'' category with 24\%, which is more than twice of the human-authored data. We are inspired that synthetic data can be manipulated and tailored for augmentation as well as adjusting the distributions.


\subsection{Human-Authored v.s. Synthetic Data}


\noindent \textbf{Discrepancy on Distributions.}
 The scatter plots in Figure~\ref{fig:dist_all} present the discrepancy between the distributions of human-authored (\textcolor{red}{red}) and synthetic dialogues (\textcolor{green}{green}), in terms of dialogue session, remediation sentence, and justification sentence respectively. We use the encoder ZH-RoBERTa\footnote{\url{https://huggingface.co/hfl/chinese-roberta-wwm-ext}} to map each context into a vector, then visualize them using T-SNE~\cite{van2008tsne}. We randomly sampled 200 human-authored dialogues and 200 synthetic dialogues which have similar dialogue scenarios. 
On the one hand, we observed salient differences in distributions between these two types of data in terms of dialogue sessions. While human-authored data is dispersed in its distribution, synthetic data is more clustered. We thus speculate that the combination of these two types of data can lead to broader and more diverse data, suitable for addressing the low-resource data condition. On the other hand, the distributions of remediation sentences generated by humans and ChatGPT are mixed with each other, demonstrating good alignment with the human capability of generating remediation measures.


\input{tables/human_compare}

\noindent \textbf{Comparison by Human Evaluation. }
To evaluate the alignment between ChatGPT and humans in terms of remediation and justification, we conduct a pair-wise comparison through human evaluation. We randomly sampled 100 utterances containing norm violations from the human-authored set and then asked ChatGPT to generate a corresponding remediation and justification sentence for each of the violation utterances. We mixed all human-authored and synthetic remediation and justification sentences and asked six annotators to judge whether each sentence met the following requirements. We mainly focus on evaluating: 1) \textit{Whether the remediation resolves the norm violation in the utterance (\textbf{Effect. of remediation})?} 2)  \textit{Whether the justification correctly explains the trigger point of the violation (\textbf{Just. on Violation})?} 3)  \textit{Whether the justification correctly explains why the remediation solves the problem (\textbf{Just. on remediation})?}
As shown in Table~\ref{tab:human_compare}, human evaluation on both human-authored and synthetic data reach high Kappa scores, indicating annotators' agreement that the quality of the remediation and justifications are high. We can observe  that the quality of synthetic data approaches human-authored ones within a small gap. These findings show the potential of ChatGPT to be aligned with human ability in the awareness of social norms.



%% file: figures/process-synthetic-collection.tex

%% file: tables/overall_statistics.tex
\begin{table}[!t]
    \scriptsize
    \centering
    \begin{tabular}{lccc}
    \toprule
       \textbf{Category}  & \textbf{Total} & \textbf{Human-authored} & \textbf{Synthetic}  \\ \midrule
      \#dialogue   & 9258 & 512 & 8746 \\
      \#utterances  & 94.36K &  7830  & 86.53K \\ 
      \#Avg. utterances & - & 15.29 & 9.90 \\
      \#violations &  21076  &   1076  &  24577\\
      \#Avg. violations & -  &  2.10 &  2.81\\ \hline
      \multicolumn{4}{l}{\#Avg. length for each following sentence} \\ \hline
      \textit{utterance} & - & 20.84 &  28.42 \\
      \textit{remediation} & - & 28.74 & 42.02 \\
      \textit{justification} & - & 35.27 & 76.64 \\
    \bottomrule
    \end{tabular}
    \caption{Statistics of \ourdata\ dataset.}
    \label{tab:dialogue_stat}
\end{table}

%% file: tables/label_statistics.tex
\begin{figure}[!t]
    \centering
    \includegraphics[width=0.33\textwidth]{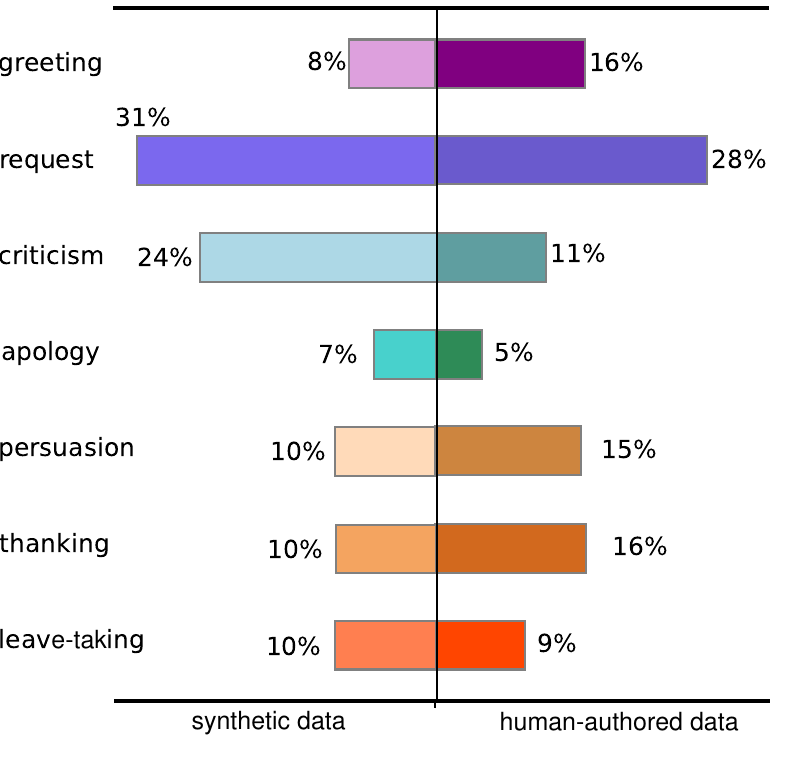}
    \caption{Norm category distributions of synthetic (\textit{left}) and human-authored (\textit{right}) data  in \ourdata\ dataset.}
    \label{fig:statistics}
\end{figure}

%% file: figures/distribution_discrepancy.tex
\begin{figure*}[t]
    \centering
     \subfigure[Dialogue session.]{
        \begin{minipage}[t]{0.31\textwidth}
        \centering
        \includegraphics[width=1\textwidth]{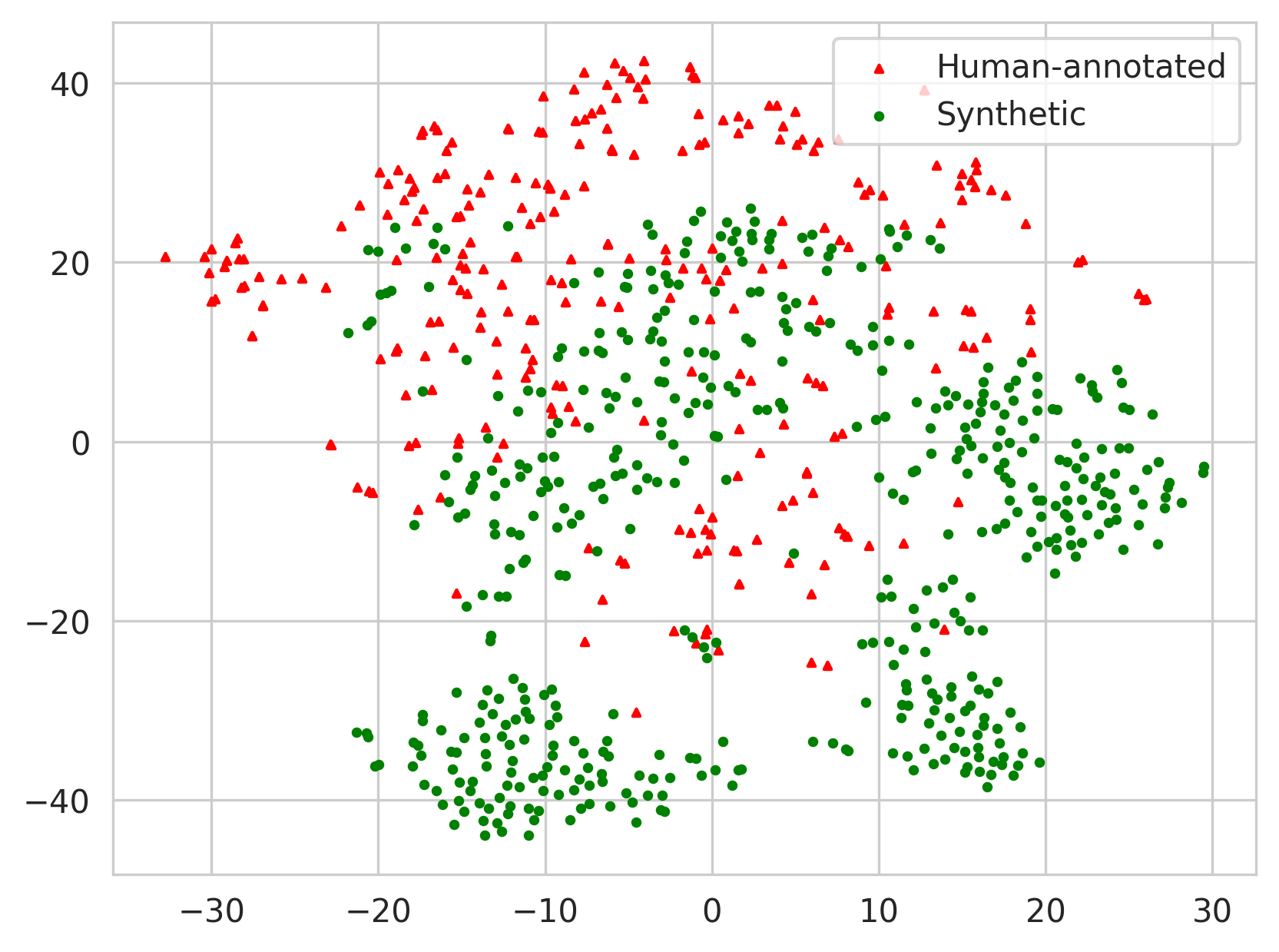}
        \label{fig:dist_dialog}
        \end{minipage}
    }
    \subfigure[Remediation sentence.]{
        \begin{minipage}[t]{0.31\textwidth}
        \centering
        \includegraphics[width=1\textwidth]{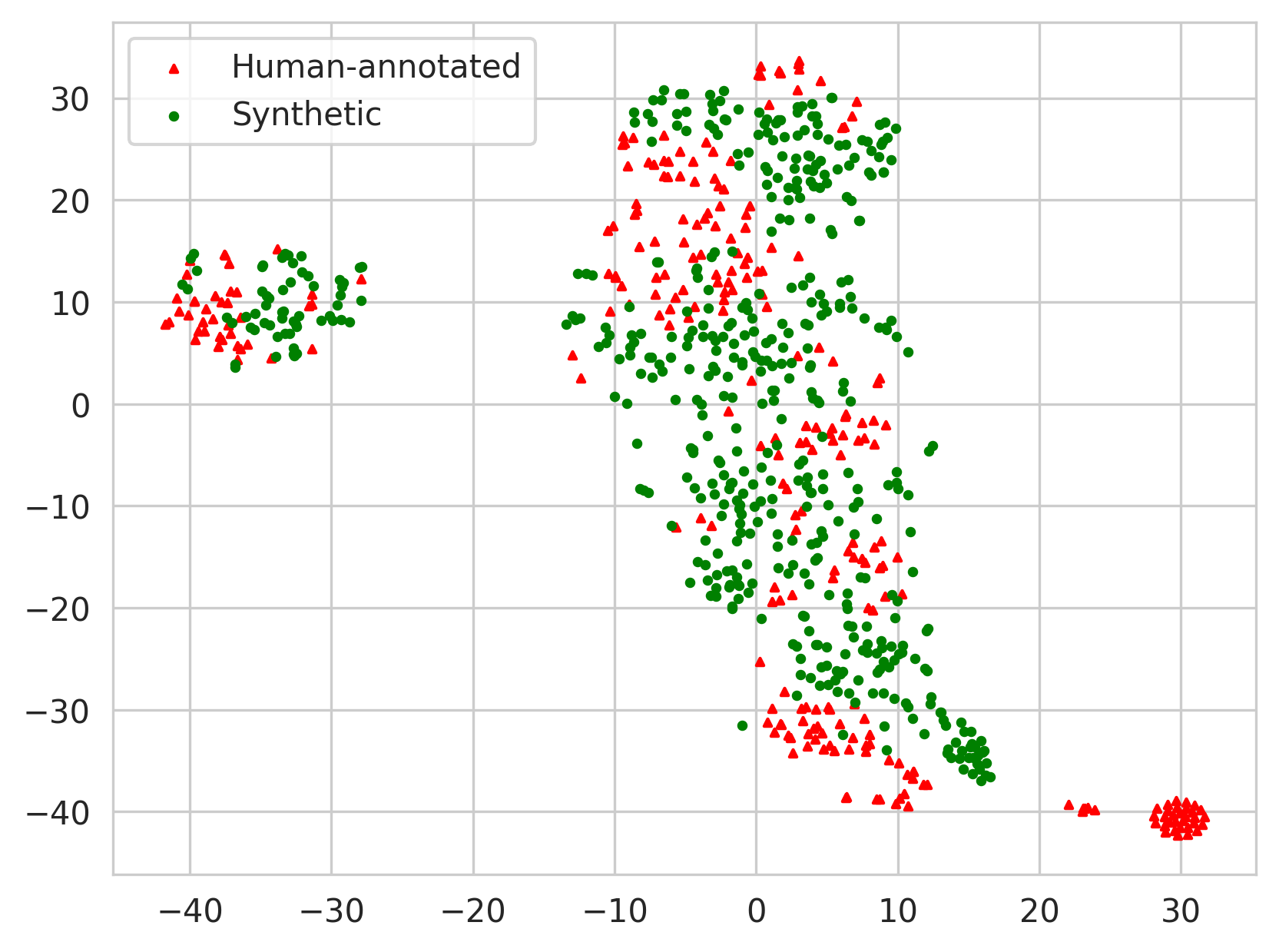}
        \label{fig:dist_reme}
        \end{minipage}
    }
    \subfigure[Justification sentence.]{
        \begin{minipage}[t]{0.31\textwidth}
        \centering
        \includegraphics[width=1\textwidth]{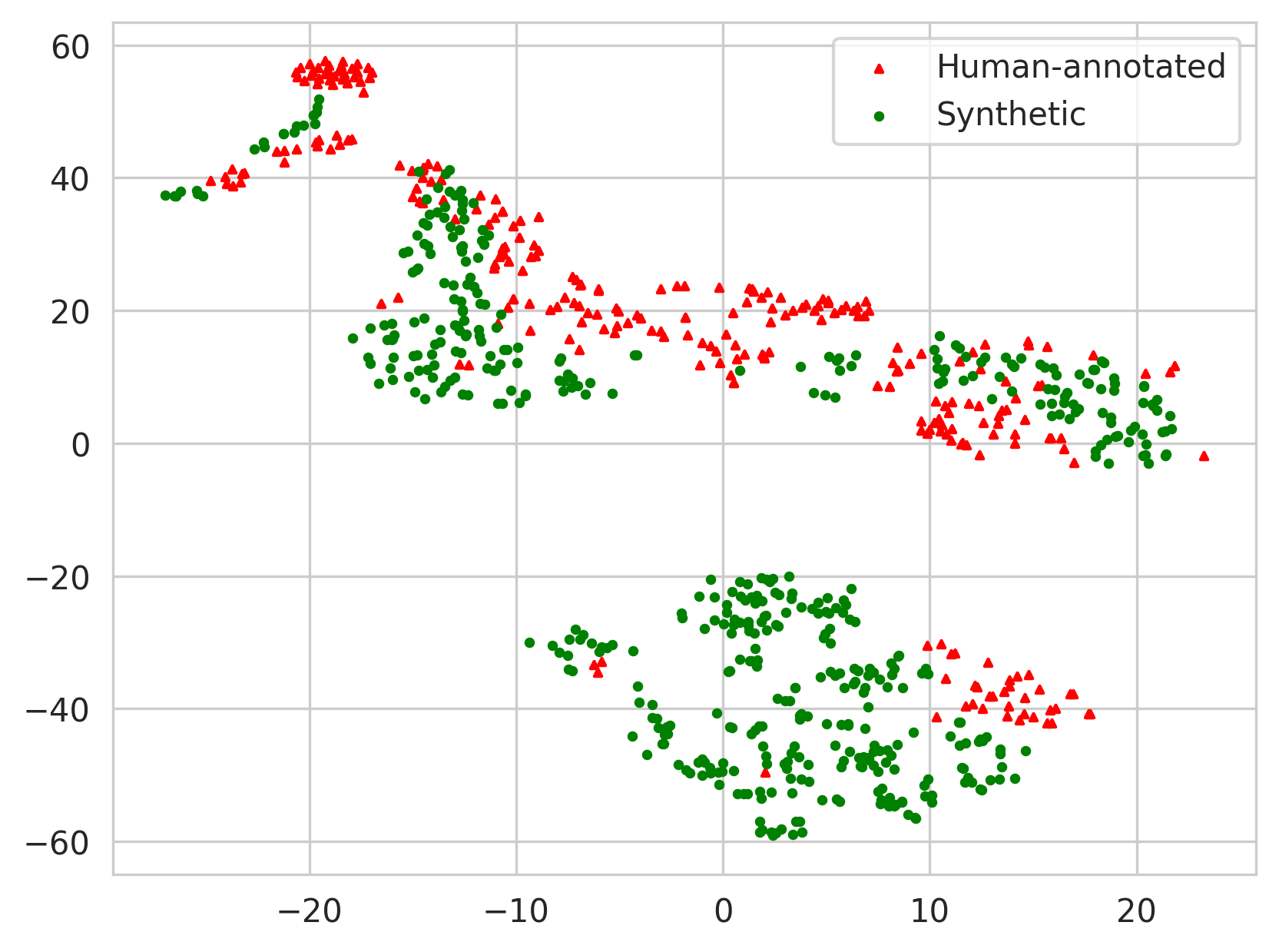}
        \label{fig:dist_just}
        \end{minipage}
    }
    \caption{Distribution divergences between the embeddings (t-SNE) of synthetic (\textcolor{green}{green}) and human-written (\textcolor{red}{red})  in terms of (a)  dialogue session, (b) remediation sentence, and (c) justification sentence.}
    \label{fig:dist_all}
\end{figure*}

%% file: tables/human_compare.tex
\begin{table}[!t]
    \scriptsize
    \centering
    \begin{tabular}{lcccccc}
    \toprule
       ~ & \multicolumn{3}{c}{Human-authored} & \multicolumn{3}{c}{Synthetic} \\ \cline{2-7}
       ~ & yes & no &  $k$ & yes & no & $k$\\ \hline

       Effect. of remediation &  96\% & 4\% &  0.79 & 87\% & 13\% &  0.63  \\
       Just. on Violation &   95\% & 5\%  & 0.73 & 90\% & 10\% & 0.68\\
       Just. on remediation & 88\% & 12\%  &  0.66 & 82\% & 18\% & 0.59\\

    \bottomrule
    \end{tabular}
    \caption{Human Comparison of human-authored and synthetic data. $k$ denotes the Cohen's Kappa score~\cite{cohen1960coefficient}.}
    \label{tab:human_compare}
\end{table}

%% file: 4-experiments.tex
\section{Experiments}

We conducted experiments to evaluate baseline performance on our proposed sub-tasks. We start with the introduction of experimental settings, followed by our analyses from the experimental results.

\subsection{Task 1: Norm Violation Detection}

\noindent \textbf{Experimental Settings.} 
 We conducted experiments to evaluate baseline performance by detecting norm categories and violations from dialogue utterances. We formulated norm category prediction as a 
 {multi-class classification task},
 while norm violation detection is a binary classification task. We used the following baseline models for our experiments: (1) \textbf{BERT-zh}: a BERT \cite{devlin-etal-2019-bert} model pre-trained on large-scale Chinese corpus. (2) \textbf{RoBERTa-zh}: a RoBERTa \cite{liu2020roberta} model pre-trained on large-scale Chinese corpus. 
 Besides, in order to explore the performance of LLMs, we employ \textbf{ChatYuan} and \textbf{ChatGPT} (3.5-turbo) as a zero-shot setting.
We organized three distinct groups, each utilizing a different source of training data: (1) exclusive training on 60\% human-authored data, (2) exclusive training on synthetic data, and (3) training on a combined dataset comprising 60\% human-authored data and all synthetic data. 
The remaining 40\% of human-authored data was divided into a validation set (10\%) and a test set (30\%) for all these settings. 
 

\input{tables/norm_category_class}

\noindent \textbf{Discussion. \textit{How LLMs perform in norm violation detection?}} 
We employed P/R/F1 scores as the evaluation metrics. The experimental results of norm category prediction and violation status detection are reported in Table~\ref{tab:norm_category_detection_results}. 
We observe that in the zero-shot setting, existing LLM (e.g., ChatYuan) only achieves 0.187 and 0.427 respectively in the norm category prediction and violation status detection tasks, which is far below the performance of fine-tuned RoBERTa-zh model. Besides, we observe that ChatGPT(3.5-turbo), the most state-of-the-art LLM, is much better than ChatYuan, but still far from good in norm violation detection task. Therefore, we urgently need a relevant corpus to benchmark LLMs or dialogue agents in aligning human interpretations of social norms.

\noindent \textbf{\textit{How synthetic data affects the performance?}}
We are curious about the necessity of synthetic data for boosting model's performance. These experiments suggest that models trained on a combination of synthetic and human data demonstrate superior performance compared to models trained solely on human data or models trained solely on synthetic data. For instance, in terms of norm category prediction,
the F1-score of BERT-zh and RoBERTa-zh trained exclusively on synthetic data is significantly inferior to the model trained on human-authored data, even though the size of synthetic training data is far greater than the human training data (8.75K $\gg$ 300). This phenomenon might be caused by the domain shift issue, as we observed a distribution gap between these two types of data in Figure~\ref{fig:dist_dialog}. However, significant improvement has been witnessed after we combined synthetic data with only a small portion of human-authored data. This finding highlights the potential of synthetic data to address data scarcity issues.

\subsection{Task 2: Violation Impact Estimation}
\noindent \textbf{Experimental Settings.}
We formulated the impact estimation task as a binary classification task. Using settings similar to Task 1, BERT-zh and RoBERTa-zh served as our baselines. We evaluated using precision, recall, and F1-score. Recognizing that prior dialogue context can influence impact estimation, we combined the current violation utterance with its two previous utterances. This combined input was then fed into our classification model.

\noindent \textbf{Discussion.} 
We present the results of violation impact estimation in Table~\ref{tab:impact_violation}. 
RoBERTa-zh model trained on a combination of synthetic and human data outperform the other two models with the other two training settings, maintaining a similar trend as the previous two tasks. However, BERT-zh model trained on the mixed data is slightly inferior than the model solely trained on human-authored data. We analyzed the bad-cases and observed that impact estimation usually requires good understanding of background culture and social norms. However, existing naive baseline models are not good enough to efficiently estimate the violation impact just from dialogue utterances. 


\input{tables/violation_impact}

\subsection{Task 3 and 4: Generation of Remediation and Justification}

\noindent \textbf{Experimental Settings.}
We employed two Chinese LLMs as the backbone models: \textbf{ChatGLM-6B}\footnote{\url{https://github.com/THUDM/ChatGLM-6B}} and \textbf{ChatYuan}\footnote{\url{https://github.com/clue-ai/ChatYuan}}, compatible with corresponding adapters: \textit{P-tuning}~\cite{ptuning2022}, \textit{Lora}~\cite{hu2022lora}, \textit{Pfeiffer}~\cite{pfeiffer2020adapterfusion} and \textit{Prefix tuning}~\cite{li2021prefix}.
Based on our investigation and results in the previous tasks, we fine-tuned the models on the combination of synthetic and human-authored training datasets and tested them on the human-authored test set.
We employ automatic evaluation metrics including BLEU~\cite{papineni2002bleu}, ROUGE-L (using F1)~\cite{lin2004rouge}, MAUVE~\cite{pillutla2021mauve}, and BERT-Score (using F1)~\cite{zhang2019bertscore}. 
Besides, we employ human evaluation to qualitatively assess the models' output by asking six human annotators to evaluate each remediation or justification sentence from three perspectives: Effectiveness (\textit{Effect.}), Relevance (\textit{Rel.}) and Informative (\textit{Info.}). Annotators are required to grade each of the remediation and justification sentences with a range of scores from 1 (low performance) to 3 (high performance).

\input{tables/auto_eval_reme_just}

\noindent \textbf{Automatic Evaluation.}
Table~\ref{tab:auto_reme} reports the automatic evaluation results on four baseline models.
We can observe that ChatYuan+\textit{Pfeiffer} achieves the best BLEU, R-L, and BScore scores and obtains the second-best score for MAUVE in the remediation generation task. This result demonstrates the strength of ChatYuan+\textit{Pfeiffer} in terms of rewriting inappropriate utterances to meet the requirements of social norms.
Besides, the remediation sentences generated by ChatYuan+\textit{Pfeiffer} have an average length of 28.78, which is very close to the human-written remediation sentences (Avg. 28.74, reported in Table~\ref{tab:dialogue_stat}). These findings demonstrate that ChatYuan+\textit{Pfeiffer} is the best among these four models to align with human capability in using concise sentences to remedy offensive utterances.

In terms of the justification generation task, ChatGLM+\textit{p-tuning} reaches the best in BLEU, MAUVE, and BScore. The generated justification sentences from ChatGLM+\textit{p-tuning} are comprehensive and detailed in illustrating the trigger point of the violation and why remediation sentences can resolve issues.
In contrast,  generated remediation and justification from the ChatGLM+\textit{Lora} model are the longest, but its performances are the lowest. We found that ChatGLM+\textit{Lora} model tends to generate tedious but irrelevant context, which cannot fulfill these two tasks in a decent format.


\noindent \textbf{Human Evaluation.}
Table~\ref{tab:human_reme} reports the annotators' manual assessments of the remediations and the corresponding justifications from three aspects.
We can observe that ChatYuan+\textit{Pfeiffer} obtains the best \textit{Rel.} and \textit{Info.} scores in the remediation task (as a reference, in Table~\ref{tab:auto_reme}, ChatYuan+\textit{Pfeiffer} ranks the first in three out of four metrics for the remediation task). Likewise, in the justification task, ChatGLM+\textit{p-tuning} performs the best in all the three human evaluation metrics, keeping the consistency with the results in Table~\ref{tab:auto_reme}.
The consistent empirical observations in both Table~\ref{tab:auto_reme} and Table~\ref{tab:human_reme} suggest that ChatYuan+\textit{Pfeiffer} can provide the best remediations to mitigate the norm violations and the ChatGLM+\textit{p-tuning} can best justify such the remediations among the four baseline models.
Also, in accordance with the finding in Table~\ref{tab:auto_reme}, ChatGLM+\textit{Lora} performs the worst in all metrics in both two tasks. This observation further verifies our previous point that verbosity probably diminishes the quality of the generated utterances. Strategically composing the outputs with more useful information and less verbosity is more important to align LLMs with humans.

\input{tables/human_eval_of_remediation_just}

\noindent \textbf{Case Study.}
Figure~\ref{fig:case_study_cn} presents a case study for the remediation and justification generation task, showcasing examples from four baseline LLMs. Among these, the ChatGLM+ptuning model excels in producing the most suitable remediation and persuasive justification. The other three models exhibit some shortcomings, such as (1) lacking politeness in remediation, and (2) containing factual inaccuracies in justification. Despite fine-tuning on our dataset, these LLMs demonstrate room for improvement in remediating violations in the future.

\input{figures/case_study_cn}

%% file: tables/norm_category_class.tex
\begin{table}[!t]
\centering
\tiny
\begin{tabular}{cc|lccc}
\toprule
\multicolumn{6}{c}{\textbf{(1) Norm Category Prediction}}             \\ \hline
 Human &  Synthetic  & \textbf{Models}    & 
 \textbf{P}   & \textbf{R}   & \textbf{F1}   \\ \hline
 \checkmark &  & BERT-zh            & 54.01        & 48.97        & 51.37         \\
 \checkmark &  & RoBERTa-zh         & 50.79        & 52.78        & 51.77         \\ \hline
  Human&  Synthetic  & \textbf{Models}    & 
 \textbf{P}   & \textbf{R}   & \textbf{F1}   \\ \hline
 &  \checkmark & BERT-zh            & 19.75        & 47.65        & 27.93         \\
 &  \checkmark & RoBERTa-zh         & 32.81        & 50.01        & 39.62      \\ \hline
  Human &  Synthetic  & \textbf{Models}    & 
 \textbf{P}   & \textbf{R}   & \textbf{F1}   \\ \hline
 \checkmark & \checkmark  & BERT-zh            & 46.64        & 82.72        & \textbf{59.65}         \\
 \checkmark &  \checkmark & RoBERTa-zh         & 48.44        & 80.76        & \textbf{60.56}         \\ \hline
\rowcolor{black!10} \multicolumn{2}{c|}{zero-shot} & ChatYuan & 12.26 & 39.57 & 18.72 \\
 \rowcolor{black!10} \multicolumn{2}{c}{setting} & GPT-3.5-turbo & 41.92 & 50.69 & 45.89 \\
 \hline \midrule

\multicolumn{6}{c}{\textbf{(2) Violation Status Detection}}  \\ \hline
 Human &  Synthetic  & \textbf{Models}    & 
 \textbf{P}   & \textbf{R}   & \textbf{F1}   \\ \hline
 \checkmark &  & BERT-zh             & 59.68        & 58.92        & 59.30         \\
 \checkmark &  & RoBERTa-zh         & 66.86        & 65.70        & 66.27       \\ \hline
  Human&  Synthetic  & \textbf{Models}    & 
 \textbf{P}   & \textbf{R}   & \textbf{F1}   \\ \hline
 &  \checkmark & BERT-zh            & 59.33        & 58.25        & 58.78       \\
 &  \checkmark & RoBERTa-zh          & 65.60        & 65.59        & 65.59     \\ \hline
  Human &  Synthetic  & \textbf{Models}    & 
 \textbf{P}   & \textbf{R}   & \textbf{F1}   \\ \hline
 \checkmark & \checkmark  & BERT-zh             & 71.04        & 68.34        & \textbf{69.66}       \\
 \checkmark &  \checkmark & RoBERTa-zh          & 67.99        & {66.97}        & \textbf{67.47}     \\ \hline
 \rowcolor{black!10} \multicolumn{2}{c|}{zero-shot} & ChatYuan & 44.68 & 40.89 & 42.70\\
 \rowcolor{black!10} \multicolumn{2}{c|}{setting} & GPT-3.5-turbo & 63.01 & 56.09 & 59.35 \\
 \hline \bottomrule
\end{tabular}%
\caption{Experiment results of Task 1 including: (1) norm category prediction and (2) violation status detection. Baseline models trained on three different settings of source data, as well as the zero-shot setting for existing two representative LLMs.}
\label{tab:norm_category_detection_results}
\end{table}

%% file: tables/violation_impact.tex
\begin{table}[!t]
\centering
\tiny
\begin{tabular}{cc|lccc}
\toprule
 Human &  Synthetic  & \textbf{Models}    & 
 \textbf{P}   & \textbf{R}   & \textbf{F1}   \\ \hline
 \checkmark &  & BERT-zh  & 74.82 & 68.97 & 71.77                   \\
 \checkmark &  & RoBERTa-zh  & 78.68 & 74.33 &  76.44           \\ \hline
  Human&  Synthetic  & \textbf{Models}    & 
 \textbf{P}   & \textbf{R}   & \textbf{F1}   \\ \hline
 &  \checkmark & BERT-zh  &  67.82 & 65.40  & 66.59          \\
 &  \checkmark & RoBERTa-zh   & 70.93 &   67.06 &  68.94        \\ \hline
  Human &  Synthetic  & \textbf{Models}    & 
 \textbf{P}   & \textbf{R}   & \textbf{F1}   \\ \hline
 \checkmark & \checkmark  & BERT-zh  & 72.48 & 69.43 & 70.92              \\
 \checkmark &  \checkmark & RoBERTa-zh  & 80.16 & 76.59 &  78.33        \\ \hline
\bottomrule
\end{tabular}%
\caption{Experiment results of impact estimation of violation models trained on three different training settings.}
\label{tab:impact_violation}
\end{table}

%% file: tables/auto_eval_reme_just.tex
\begin{table}[!t]
    \centering
    \tiny
    \begin{tabular}{lccccc}
    \toprule
      \multicolumn{6}{c}{\textbf{Remediation Generation}} \\ 
       Model  &  BLEU. & R-L & MAUVE & BScore & Avg. Len\\ \hline
       ChatGLM + \textit{P-tuning}  &  0.211 & 0.308 & \textbf{0.598} & 0.694 & 38.73    \\
        ChatGLM + \textit{Lora}  &  0.129 & 0.161 & 0.005 & 0.610 & {213.38}   \\
        ChatYuan + \textit{Pfeiffer}  & \textbf{0.244} & \textbf{0.359} & 0.384 & \textbf{0.713} & 28.78    \\
          ChatYuan + \textit{Prefix tuning}  & 0.161 & 0.311 & 0.280 & 0.699 & 17.93    \\ \hline\midrule
         \multicolumn{6}{c}{\textbf{Justification Reason Generation}} \\ 
      Model  &  BLEU. & R-L & MAUVE & BScore & Avg. Len\\ \hline
       ChatGLM + \textit{P-tuning}  & \textbf{0.117} & 0.144 & \textbf{0.025} & \textbf{0.612} & 93.21 \\
        ChatGLM + \textit{Lora}  &   0.085 & 0.082 & 0.005 & 0.554 & {244.05} \\
        ChatYuan + \textit{Pfeiffer}  & {0.106} & 0.150 & 0.014 & 0.603 & 66.46     \\
          ChatYuan + \textit{Prefix tuning}  & 0.103 & \textbf{0.154} & 0.014 & 0.611 & 58.10   \\
    \bottomrule
    \end{tabular}
    \caption{Automatic evaluation on the remediation generation and justification reason generation task respectively.}
    \label{tab:auto_reme}
\vspace{-3mm}
\end{table}

%% file: tables/human_eval_of_remediation_just.tex
\begin{table}[!t]
    \centering
    \tiny
    \begin{tabular}{lcccc}
    \toprule
      \multicolumn{5}{c}{\textbf{Remediation Generation}} \\ 
       Model  &  Effect. & Rel. & Info. & \textit{kappa} \\ \hline
       ChatGLM + \textit{P-tuning}  & \textbf{2.33}	& 2.42	& 2.36 & 0.53  \\
        ChatGLM + \textit{Lora}  &  1.37 & 1.62 & 1.39 & 0.61   \\
        ChatYuan + \textit{Pfeiffer}  &  2.29 & 	\textbf{2.71} & 	\textbf{2.79} &  0.49   \\
          ChatYuan + \textit{Prefix tuning}  & 1.94 & 2.35 & 2.16 &   0.56  \\ \hline\midrule
         \multicolumn{5}{c}{\textbf{Justification Reason Generation}} \\ 
       Model  &  Effect. & Rel. & Info. & \textit{kappa} \\ \hline
       ChatGLM + \textit{P-tuning}  & \textbf{2.65} & \textbf{2.71} & \textbf{2.76}   & 0.55  \\
        ChatGLM + \textit{Lora}  &  1.83 & 2.32 & 2.20  & 0.59 \\
        ChatYuan + \textit{Pfeiffer}  &  2.46 & 2.59 & 2.72 & 0.55  \\
          ChatYuan + \textit{Prefix tuning}  & 2.14 & 2.48 & 2.25   & 0.57 \\
    \bottomrule
    \end{tabular}
    \caption{Human evaluation on the remediation generation and justification reason generation task respectively.}
    \label{tab:human_reme}
\end{table}

%% file: figures/case_study_cn.tex
\begin{figure}[!t]
    \centering
    \includegraphics[width=0.5\textwidth]{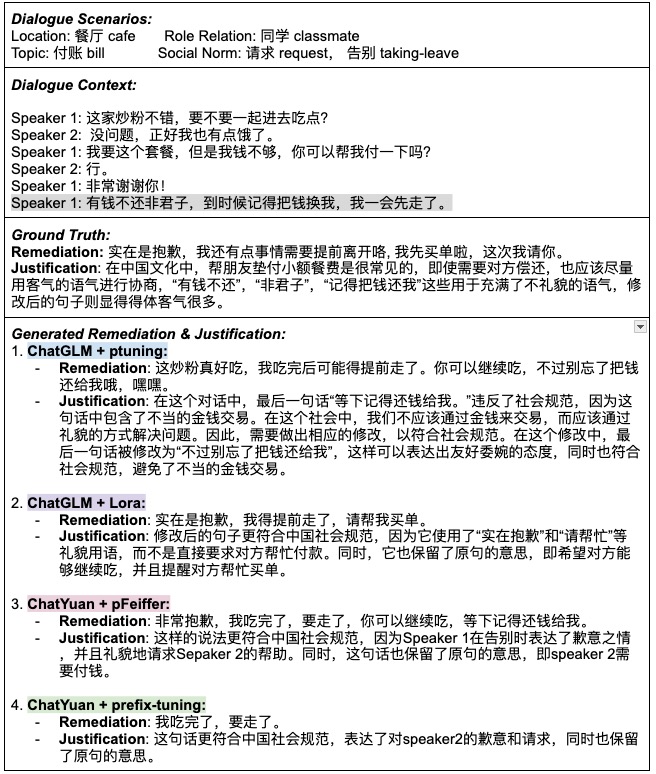}
    \caption{Case study on the generated remediation and justification from different LLMs. 
    \colorbox{black!10}{We refer the corresponding English translation version}  \colorbox{black!10}{to appendix~\ref{sec:case_study_en}.}}
    \label{fig:case_study_cn}
\end{figure}

%% file: 5-relatedwork.tex
\section{Related Work}
\label{sec:relatedwork}

\paragraph{Socio-Cultural Conversations.}
Culture refers to shared values, behaviors and beliefs within a society~\cite{lytle1995paradigm}. These shared values, behaviors and beliefs give rise to social norms, which are an implicitly held consensus on social patterns governing behavior within a cultural context~\cite{morris2015normology,hua2024sadas}.
Social norms in conversations have been recognized as an important factor that affects human's communication~\cite{chawla-etal-2023-social}, such as negotiation~\cite{chen2023put,zhan2024let} or dialects~\cite{joshi2024natural}.
\citet{forbes-etal-2020-social} propose a large-scale corpus
-- \textit{Social-Chemistry-101}, containing 292K rules-of-thumb (RoTs); \citet{hendrycks2021aligning} introduce the ETHICS dataset,
where the task is to predict moral judgments about diverse scenarios; and \citet{ziems-etal-2022-moral} propose a moral-related
corpus using 99K distinct RoTs to explore ethical issues in dialogue. These datasets are formulated as single context-response
pairs, hence they do not simulate real dialogues.
When observed behaviors do not conform to what is expected, norm violations occur, which may lead to potential conflicts~\cite{burgoon1993interpersonal}. 
{\citet{zhan2023socialdial} propose a corpus to detect norm violations in multi-turn conversations, but falling short of resolving the violation issues.}
Moreover, remediation tactics that transform a negative impression caused by norm violations to a positive one are essential to benchmark LLM's ability to mitigate potential harm. 
To the best of our knowledge, our \ourdata\ dataset is the first corpus to explore how to remediate norm violations in socio-cultural conversations.

\paragraph{Synthetic Data for Dialogues.} 
Synthetic data is regarded as an effective approach to accommodate data scarcity for low-resource dialogue systems~\cite{zhan2023turning}.
\citet{dai2022dialog} propose a novel task called \textit{Dialogue Inpainting}, which transforms an input raw document into a two-party QA session.
In addition, the emergence of large language models (LLMs) has greatly advanced many NLP tasks. In terms of synthetic data for dialogue,
\citet{kim2022botstalk} propose a framework for automatic curation of large-scale multi-skill dialogue datasets;
\citet{chen-etal-2023-places} utilize LLMs and devise a prompt-based framework to create synthetic conversations for few-shot social dialogue.
{Compared with existing methods, the synthetic dialogues in \ourdata\ are generated with ChatGPT, which can be used for augmenting low-resource settings, as well as assessing the alignment between LLMs and humans.
}

%% file: 6-conclusion.tex
\section{Conclusion}

We propose \ourdata, a Chinese socio-cultural conversation benchmark, to explore how to remediate norm violations. \ourdata\ contains 9,258 dialogue sessions in total, of which 512 dialogues are written by humans and 8,746 synthetic dialogues are generated by ChatGPT. To the best of our knowledge, \ourdata\ is the first multi-turn dialogue corpus to study norm violation remediation in conversations. 
Based on the EVT and IAT theories, we formulate four tasks to help understand, detect and remediate social norm violations.
We further conduct in-depth analyses on these sub-tasks in succession and assessed several popular LLMs' performances.

%% file: 7-limitations.tex
\section*{Limitations}

We claim that our work may have limitations in the following aspects.

\paragraph{Monolingual Culture Background}
As a pioneer work for norm violation remediation in dialogues, we mainly focus on Chinese social norms and offer a Chinese dataset. In future, we will extend our dataset to a cross-cultural and multilingual corpus, which will involve more culture backgrounds, such as Spanish, Latin and Arabic.

\paragraph{Lack of Tailored Baseline Models}
We are aware of that our work is the first to propose norm violation remediation and justification tasks. Our contributions mainly focus on formulating relevant tasks and datasets, thus falling short on proposing tailored baseline models.

%% file: 8-Ethics.tex
\section*{Ethics Statement}

To regularize the usage of this resource and the tasks it can facilitate, we will claim several ethics consideration and emphasize some potential risks.

\paragraph{Misuse of Data.}
As the objective of this resource is to integrate AI systems with the remediation ability towards norm violations. Inevitably, this resource will contain some content that may be offensive or upsetting. However, we want to stress that \ourdata\ represents a collection of social norm, remediation and justification. We do not treat the norm violations as discrimination, racism or disrespect to Chinese or any other cultures.
Therefore, this dataset, primarily synthesized using LLMs and crowd-sourced inputs, is released exclusively for academic and research purposes and does not reflect the opinions or values of the authors. The social norms and violation situations in \ourdata\ are strictly prohibited for any form of commercial exploitation or political manipulation. They should not be used as insults, slander, or for other malicious intents. Users are expected to adhere to the highest ethical standards, ensuring responsible and transparent use aligned with ethical research practices. The dataset creators hold no liability for misuse or misinterpretation, and all necessary measures have been taken to respect privacy and ensure informed consent in the data collection process.

\paragraph{Risks in Annotation.}
We highly value our annotators' mental health and labor compensation. Before and annotation, data collection or human evaluation, relevant studies were carefully reviewed and approved by an internal review board. Our task may contain some offensive or upsetting content. We thus require each annotator to have a rest every one hour or anytime they do not feel well. In terms of payment, we pay these annotators 15 USD/hour.

%% file: 9-appendix.tex
\section{Corresponding English translation of Figure~\ref{fig:case_study_cn}}
\label{sec:case_study_en}

We put the corresponding English translation of Figure~\ref{fig:case_study_cn} in the Figure~\ref{fig:case_study_en}.

\input{figures/case_study_en}

\section{Details of \ourdata\ dataset}

\input{tables/social_norm_definition}

\subsection{Definition of Social Norms}
\label{sec:def-social-norm}

We present the catgories of social norms and corresponding rules and examples in Table~\ref{tab:social_norm_definition}, which covers all of the types that appear in our human-authored and synthetic dataset. We mainly focus on seven norm categories in our paper, including: \textit{apology}, \textit{criticism}, \textit{greeting}, \textit{persuasion}, \textit{request}, \textit{leave-taking} and \textit{thanking}. We notice that these norm categories may have overlaps with the definition of dialogue acts~\cite{stolcke2000dialogue} or intents~\cite{wen2017latent}. However, we want to stress that the main difference of the rules on social norms relies on: \textbf{socially or culturally accepted behaviors} within these actions/norms.

\subsection{Details of Taxonomy in Dialogue Scenarios}

We present the relevant social factors including \textit{location} and \textit{role relation} in Figure~\ref{fig:social_factor}. Some other keywords in low frequency are not presented in this table. Each dialogues in both human-authored and synthetic dataset will contain a value for each social factors.

\input{figures/social_factors}

\section{Example of Synthetic Data Generation}
\label{synthetic_ex_app}

We present a example for the synthetic data generation procedure in Figure~\ref{fig:synthetic_ex}. We devise a ontology-based framework to gradually prompt ChatGPT to generate synthetic conversations. Overall, three main steps included in the ontology-based framework: (1) Norm Violation Example Generation, (2) Synthetic Conversation Generation and (3) Remediation and Justification Generation. Specifically, Step 1 Norm Violation Example Generation will generate basic norm rules and several violation examples in Chinese culture considering the provided dialogue scenarios above. Step 2 Synthetic Conversation Generation will generate synthetic dialogues that contain above mentioned violation examples. Additionally, corresponding labels such as norm category, violation status will be annotated automatically.
Based on these utterances which contains norm violation, Step 3 will generation corresponding remediation sentence and justification sentence as shown in Figure~\ref{fig:synthetic_ex}.

\input{figures/synthetic_example}

\section{Comparison of \ourdata and other datasets}

We present the statistical comparision between \ourdata\ and other relevant datasets. \ourdata\  
differs from previous datasets in the following aspects: (1) to the best of our knowledge, it is the first dataset aiming at remediating the social norm violations
based on Chinese social norms, and \ourdata\ covers at most seven different social norm categories; (2) besides norm violation detection task, we firstly define the norm violation remediation and justification task, and collect high-quality human-authored and automatically generated synthetic data from ChatGPT, which provides the benchmark to assess the alignment between human and LLMs in awareness of social norms.

\begin{table*}[ht]
\begin{adjustbox}{ width=\textwidth,center}
\centering
\begin{tabular}{lccccccc}
\hline
\textbf{Dataset}   & \textbf{Type} &  \textbf{\#Dialogues} & \textbf{\#Avg. turns} & \textbf{language} & \textbf{social factors} & \textbf{Remediation of Norm Violations} & \textbf{Latest Updates }\\
\hline
FactAct~\cite{dutt2020keeping} & multi-turn &   299 & 35.8  & English & persuasion & \XSolidBrush & 2018 \\
PersuasionforGood~\cite{wang2019persuasion} & multi-turn & 1017 & 10.43  & English & request, persuasion & \XSolidBrush & 2019\\
CPED~\cite{chen2022cped} & multi-turn & 12k & 11.08 & Chinese & emotion & \XSolidBrush & 2022  \\
moralInt~\cite{ziems2022moralIntegrityCorpus} & single-turn &  38k & - & English & norm rule & \XSolidBrush & 2022 \\
DREAM~\cite{gu2022dream} & single-turn & 49k & - & English & norm rule & \XSolidBrush  & 2022 \\ 
SocialDial~\cite{zhan2023socialdial}  & multi-turn   & 6433 & 9.45 & Chinese  & norm rule & \XSolidBrush & 2023 \\ \hline
\ourdata & multi-turn & 9,258 & 10.19 & Chinese & norm rule & \CheckmarkBold & 2023 \\

\bottomrule
\end{tabular}
\vspace*{2mm}
\end{adjustbox}
\caption{Comparison between \ourdata\ and related dialogue corpora.}
\label{tab:compare}
\vspace*{-6mm}
\end{table*}

%% file: figures/case_study_en.tex
\begin{figure}[!t]
    \centering
    \includegraphics[width=0.5\textwidth]{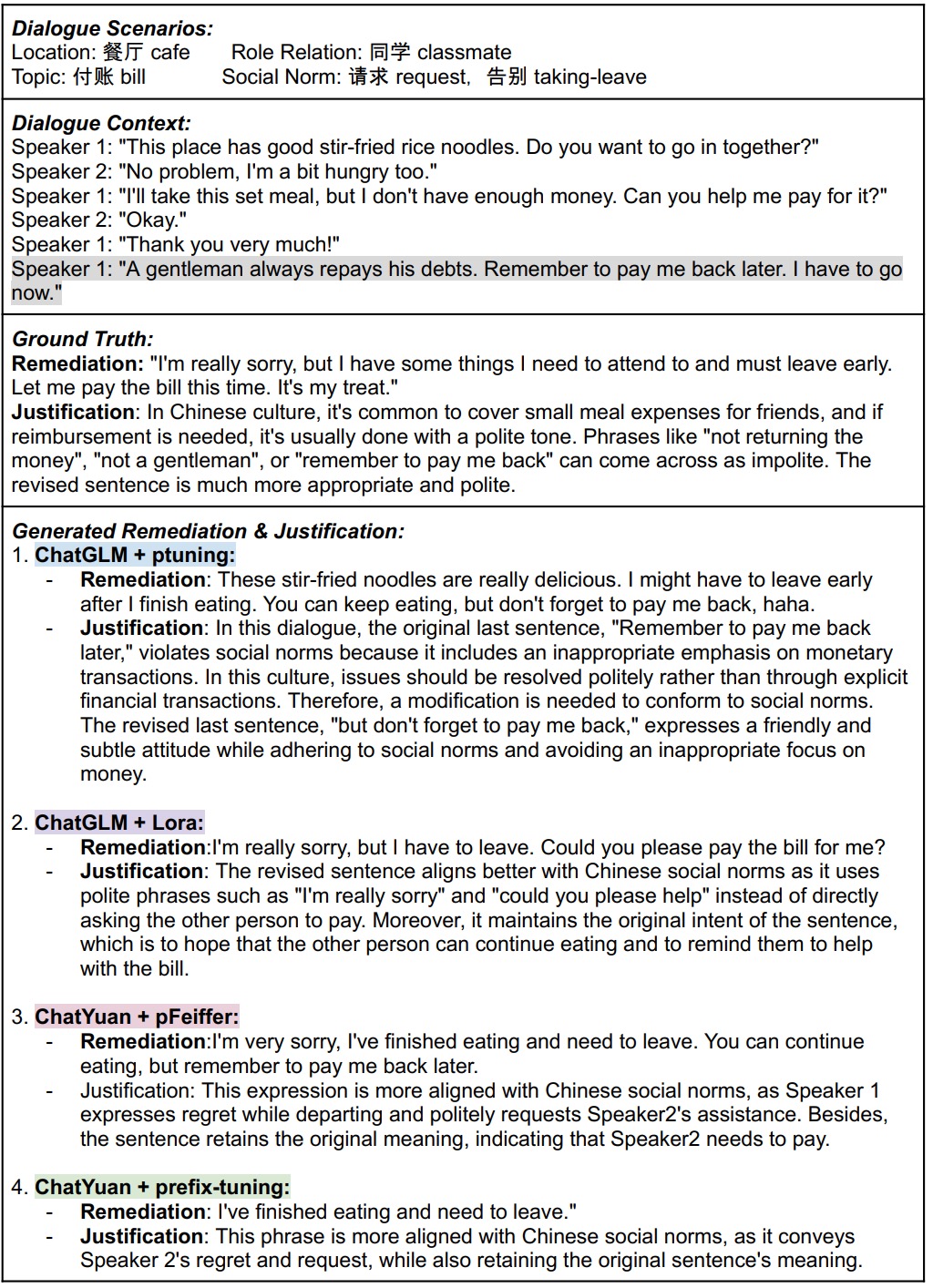}
    \caption{The English verson of case study on the generated remediation and justification from different LLMs. Please note that the translation was conducted by ChatGPT.}
    \label{fig:case_study_en}
\end{figure}

%% file: tables/social_norm_definition.tex
\begin{table*}[!th]
\centering
\resizebox{\linewidth}{!}{%
\begin{tabular}{llll}
\hline
\# & Norm Category & Norm Rules                                                                                                                                                                                                                                                                              & Examples                                                                                      \\ \hline
1 & Apology       & \begin{tabular}[c]{@{}l@{}}Apologies in Mandarin/Chinese culture are guided by principles of harmony as well as honor, \\ dignity and respect. Direct verbal apologies might be avoided when an indirect approach, such \\ as offering a wordless gesture or a written message, can be taken instead.\end{tabular}    & \begin{tabular}[c]{@{}l@{}} \begin{CJK}{UTF8}{gbsn}我向你道歉 。\end{CJK}\\ (I apologize to you.)\end{tabular} \\

2 & Criticism     & \begin{tabular}[c]{@{}l@{}}In Chinese culture it is common for direct criticism to be given to subordinates or those 
of \\ lower status, while criticizing a superior or someone of higher status is uncommon and is \\typically done in a much more indirect manner.\end{tabular}   & \begin{tabular}[c]{@{}l@{}}\begin{CJK}{UTF8}{gbsn}1. 上级对下级：你这么做的方式不太对。\end{CJK}\\ (What you're doing is not totally correct.)    \\ \begin{CJK}{UTF8}{gbsn}2. 下级对上级：部长先生，这里似乎有一个错别字需要改正一下，您看呢？\end{CJK}\\ (Mr. Minister, there seems to be a typo here that needs correction. am I right?)\end{tabular} \\

3 & Greeting      & \begin{tabular}[c]{@{}l@{}}Using specific greetings in Mandarin Chinese culture is very important in formal settings. \\ However, greetings are far more relaxed in intimate relationships such as with  family \\ and friends or people of similar or
younger ages and in informal settings.  \end{tabular} & \begin{tabular}[c]{@{}l@{}} \begin{CJK}{UTF8}{gbsn} 1. 部长先生，早上好，很高兴见到你！\end{CJK}\\ (Mr. Minister, good morning, it's a pleasure to see you!)\\ \begin{CJK}{UTF8}{gbsn}2. 早哟！\end{CJK}\\ (Morning!)\end{tabular}        \\
4 & Persuasion    & \begin{tabular}[c]{@{}l@{}}In Mandarin Chinese culture, the norm of doing persuasion varies by speakers' social
status \\ and age. persuasion involves people giving reasons and/or describing consequences if things \\are done one way or the other.\end{tabular}                                                                                             & \begin{tabular}[c]{@{}l@{}} \begin{CJK}{UTF8}{gbsn}建议你可以... \end{CJK}\\ (I suggest you ....)\end{tabular}                      \\
5 & Request       & \begin{tabular}[c]{@{}l@{}}In Chinese culture, factors such as status, power, age, gender, and familiarity play a large  role \\ in determining the way in which requests are made. it's preferable to use a politeness marker. \end{tabular}                                                                                & \begin{tabular}[c]{@{}l@{}}\begin{CJK}{UTF8}{gbsn}请问你有时间帮我做...吗?\end{CJK}\\ (May I ask if you have the time to...?)\end{tabular}          \\
6 & Leave-taking  & \begin{tabular}[c]{@{}l@{}}In Mandarin Chinese culture, taking leave is a multi-stage process, and social norms around \\ taking leave vary by social status, age. The person who is taking leave usually starts with \\apologizing or giving a reason or an excuse for leaving.\end{tabular}                                                                                                                                         & \begin{tabular}[c]{@{}l@{}}\begin{CJK}{UTF8}{gbsn}实在抱歉，我后面还有个安排，今天的会就到这吧!\end{CJK}\\ (Sorry guys, I have another schedule afterwards, Let's end the meeting today.)\end{tabular}                  \\
7 & Thanking        & \begin{tabular}[c]{@{}l@{}}Thanking people directly in Mandarin Chinese culture is frequent in formal settings or\\
when interacting with people of a higher status or equal status. The norm of doing thanks \\ should expresses gratitude to a person, or institution.\end{tabular}                                                                                                                                  & \begin{tabular}[c]{@{}l@{}}\begin{CJK}{UTF8}{gbsn}太谢谢了!\end{CJK}\\ (Thank you very much!)\end{tabular}                        \\
8 & Others        & Other norms that are not included in the previous categories.                                                                                                                                                                                                                           &                                                                                               \\ \hline
\end{tabular}%
}
\caption{Social norm categories and corresponding rules in our main seven categories.}
\label{tab:social_norm_definition}
\end{table*}

%% file: figures/social_factors.tex
\begin{figure}[!t]
    \centering
    \includegraphics[width=0.5\textwidth]{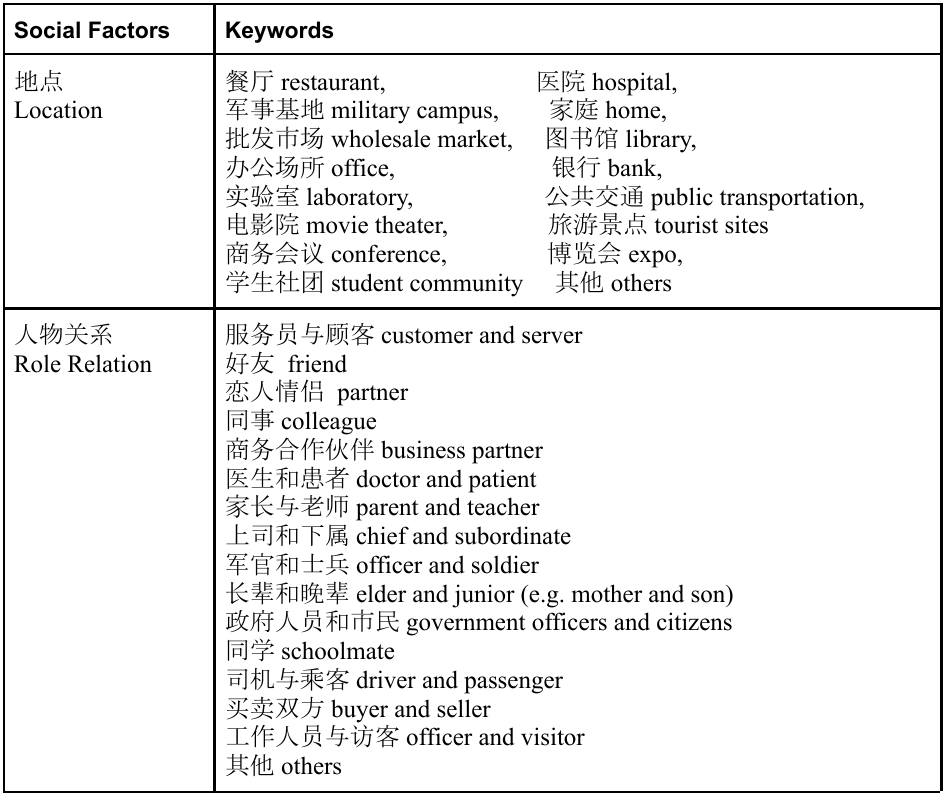}
    \caption{Taxonomy of social factors in dialogue scenarios.}
    \label{fig:social_factor}
\end{figure}

%% file: figures/synthetic_example.tex
\begin{figure}[!t]
    \centering
    \includegraphics[width=0.5\textwidth]{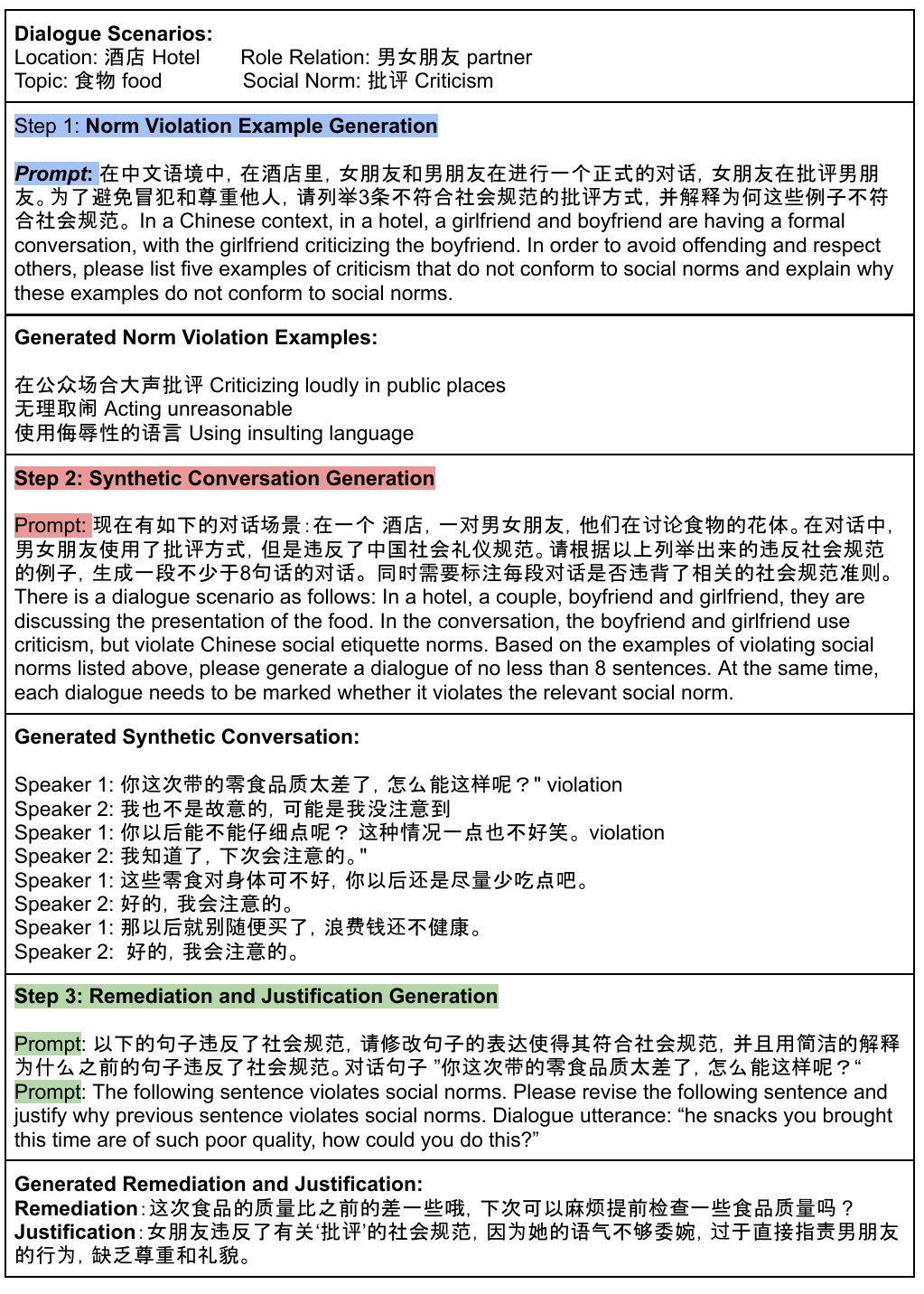}
    \caption{A example of generating synthetic conversation by prompting ChatGPT with three steps.}
    \label{fig:synthetic_ex}
\end{figure}

%% file: Main.bbl
\begin{thebibliography}{48}
\expandafter\ifx\csname natexlab\endcsname\relax\def\natexlab#1{#1}\fi

\bibitem[{Bicchieri et~al.(2018)Bicchieri, Muldoon, and Sontuoso}]{sep-social-norms}
Cristina Bicchieri, Ryan Muldoon, and Alessandro Sontuoso. 2018.
\newblock {Social Norms}.
\newblock In Edward~N. Zalta, editor, \emph{The {Stanford} Encyclopedia of Philosophy}, {W}inter 2018 edition. Metaphysics Research Lab, Stanford University.

\bibitem[{Burgoon(1993)}]{burgoon1993interpersonal}
Judee~K Burgoon. 1993.
\newblock Interpersonal expectations, expectancy violations, and emotional communication.
\newblock \emph{Journal of language and social psychology}, 12(1-2):30--48.

\bibitem[{Burgoon(2015)}]{burgoon2015expectancy}
Judee~K Burgoon. 2015.
\newblock Expectancy violations theory.
\newblock \emph{The international encyclopedia of interpersonal communication}, pages 1--9.

\bibitem[{Burgoon and Hubbard(2005)}]{burgoon2005cross}
Judee~K Burgoon and AS~Ebesu Hubbard. 2005.
\newblock Cross-cultural and intercultural applications of expectancy violations theory and interaction adaptation theory.
\newblock \emph{Theorizing about intercultural communication}, pages 149--171.

\bibitem[{Burgoon and Walther(1990)}]{burgoon1990nonverbal}
Judee~K Burgoon and Joseph~B Walther. 1990.
\newblock Nonverbal expectancies and the evaluative consequences of violations.
\newblock \emph{Human Communication Research}, 17(2):232--265.

\bibitem[{Burgoon et~al.(1997)Burgoon, White, and Greene}]{burgoon1997researching}
Judee~K Burgoon, Cindy~H White, and John~O Greene. 1997.
\newblock Researching nonverbal message production: A view from interaction adaptation theory.
\newblock \emph{Message production: Advances in communication theory}, pages 279--312.

\bibitem[{Chawla et~al.(2023)Chawla, Shi, Zhang, Lucas, Yu, and Gratch}]{chawla-etal-2023-social}
Kushal Chawla, Weiyan Shi, Jingwen Zhang, Gale Lucas, Zhou Yu, and Jonathan Gratch. 2023.
\newblock \href {https://aclanthology.org/2023.eacl-main.53} {Social influence dialogue systems: A survey of datasets and models for social influence tasks}.
\newblock In \emph{Proceedings of the 17th Conference of the European Chapter of the Association for Computational Linguistics}, pages 750--766, Dubrovnik, Croatia. Association for Computational Linguistics.

\bibitem[{Chen et~al.(2023{\natexlab{a}})Chen, Yuan, Ye, Majumder, and Richardson}]{chen2023put}
Jiangjie Chen, Siyu Yuan, Rong Ye, Bodhisattwa~Prasad Majumder, and Kyle Richardson. 2023{\natexlab{a}}.
\newblock Put your money where your mouth is: Evaluating strategic planning and execution of llm agents in an auction arena.
\newblock \emph{arXiv preprint arXiv:2310.05746}.

\bibitem[{Chen et~al.(2023{\natexlab{b}})Chen, Papangelis, Tao, Kim, Rosenbaum, Liu, Yu, and Hakkani-Tur}]{chen-etal-2023-places}
Maximillian Chen, Alexandros Papangelis, Chenyang Tao, Seokhwan Kim, Andy Rosenbaum, Yang Liu, Zhou Yu, and Dilek Hakkani-Tur. 2023{\natexlab{b}}.
\newblock \href {https://aclanthology.org/2023.findings-eacl.63} {{PLACES}: Prompting language models for social conversation synthesis}.
\newblock In \emph{Findings of the Association for Computational Linguistics: EACL 2023}, pages 844--868, Dubrovnik, Croatia. Association for Computational Linguistics.

\bibitem[{Chen et~al.(2022)Chen, Fan, Xing, Pang, Huang, Han, Tie, and Xu}]{chen2022cped}
Yirong Chen, Weiquan Fan, Xiaofen Xing, Jianxin Pang, Minlie Huang, Wenjing Han, Qianfeng Tie, and Xiangmin Xu. 2022.
\newblock Cped: A large-scale chinese personalized and emotional dialogue dataset for conversational ai.
\newblock \emph{arXiv preprint arXiv:2205.14727}.

\bibitem[{Chiu et~al.(2010)Chiu, Gelfand, Yamagishi, Shteynberg, and Wan}]{chiu2010intersubjective}
Chi-Yue Chiu, Michele~J Gelfand, Toshio Yamagishi, Garriy Shteynberg, and Ching Wan. 2010.
\newblock Intersubjective culture: The role of intersubjective perceptions in cross-cultural research.
\newblock \emph{Perspectives on Psychological Science}, 5(4):482--493.

\bibitem[{Cohen(1960)}]{cohen1960coefficient}
Jacob Cohen. 1960.
\newblock A coefficient of agreement for nominal scales.
\newblock \emph{Educational and psychological measurement}, 20(1):37--46.

\bibitem[{Dai et~al.(2022)Dai, Chaganty, Zhao, Amini, Rashid, Green, and Guu}]{dai2022dialog}
Zhuyun Dai, Arun~Tejasvi Chaganty, Vincent~Y Zhao, Aida Amini, Qazi~Mamunur Rashid, Mike Green, and Kelvin Guu. 2022.
\newblock Dialog inpainting: Turning documents into dialogs.
\newblock In \emph{International Conference on Machine Learning}, pages 4558--4586. PMLR.

\bibitem[{Devlin et~al.(2019)Devlin, Chang, Lee, and Toutanova}]{devlin-etal-2019-bert}
Jacob Devlin, Ming-Wei Chang, Kenton Lee, and Kristina Toutanova. 2019.
\newblock \href {https://doi.org/10.18653/v1/N19-1423} {{BERT}: Pre-training of deep bidirectional transformers for language understanding}.
\newblock In \emph{Proceedings of the 2019 Conference of the North {A}merican Chapter of the Association for Computational Linguistics: Human Language Technologies, Volume 1 (Long and Short Papers)}, pages 4171--4186, Minneapolis, Minnesota. Association for Computational Linguistics.

\bibitem[{Dutt et~al.(2020)Dutt, Joshi, and Ros{\'e}}]{dutt2020keeping}
Ritam Dutt, Rishabh Joshi, and Carolyn~Penstein Ros{\'e}. 2020.
\newblock Keeping up appearances: Computational modeling of face acts in persuasion oriented discussions.
\newblock \emph{arXiv preprint arXiv:2009.10815}.

\bibitem[{Ebesu~Hubbard(2015)}]{ebesu2015IAT}
Amy~S Ebesu~Hubbard. 2015.
\newblock Interaction adaptation theory.
\newblock \emph{The International Encyclopedia of Interpersonal Communication}, pages 1--5.

\bibitem[{Feng et~al.(2023)Feng, Qu, and Haffari}]{10.1162/tacl_a_00561}
Tao Feng, Lizhen Qu, and Gholamreza Haffari. 2023.
\newblock \href {https://doi.org/10.1162/tacl_a_00561} {{Less is More: Mitigate Spurious Correlations for Open-Domain Dialogue Response Generation Models by Causal Discovery}}.
\newblock \emph{Transactions of the Association for Computational Linguistics}, 11:511--530.

\bibitem[{Forbes et~al.(2020{\natexlab{a}})Forbes, Hwang, Shwartz, Sap, and Choi}]{forbes2020social}
Maxwell Forbes, Jena~D Hwang, Vered Shwartz, Maarten Sap, and Yejin Choi. 2020{\natexlab{a}}.
\newblock Social chemistry 101: Learning to reason about social and moral norms.
\newblock In \emph{Proceedings of the 2020 Conference on Empirical Methods in Natural Language Processing (EMNLP)}, pages 653--670.

\bibitem[{Forbes et~al.(2020{\natexlab{b}})Forbes, Hwang, Shwartz, Sap, and Choi}]{forbes-etal-2020-social}
Maxwell Forbes, Jena~D. Hwang, Vered Shwartz, Maarten Sap, and Yejin Choi. 2020{\natexlab{b}}.
\newblock \href {https://doi.org/10.18653/v1/2020.emnlp-main.48} {Social chemistry 101: Learning to reason about social and moral norms}.
\newblock In \emph{Proceedings of the 2020 Conference on Empirical Methods in Natural Language Processing (EMNLP)}, pages 653--670, Online. Association for Computational Linguistics.

\bibitem[{Fung et~al.(2022)Fung, Chakraborty, Guo, Rambow, Muresan, and Ji}]{fung2022normsage}
Yi~R Fung, Tuhin Chakraborty, Hao Guo, Owen Rambow, Smaranda Muresan, and Heng Ji. 2022.
\newblock Normsage: Multi-lingual multi-cultural norm discovery from conversations on-the-fly.
\newblock \emph{arXiv preprint arXiv:2210.08604}.

\bibitem[{Gu et~al.(2022)Gu, Dalvi, and Clark}]{gu2022dream}
Yuling Gu, Bhavana Dalvi, and Peter Clark. 2022.
\newblock Dream: Improving situational qa by first elaborating the situation.
\newblock In \emph{Proceedings of the 2022 Conference of the North American Chapter of the Association for Computational Linguistics: Human Language Technologies}, pages 1115--1127.

\bibitem[{Hendrycks et~al.(2021)Hendrycks, Burns, Basart, Critch, Li, Song, and Steinhardt}]{hendrycks2021aligning}
Dan Hendrycks, Collin Burns, Steven Basart, Andrew Critch, Jerry Li, Dawn Song, and Jacob Steinhardt. 2021.
\newblock Aligning ai with shared human values.
\newblock \emph{ICLR}.

\bibitem[{Hu et~al.(2022)Hu, Shen, Wallis, Allen-Zhu, Li, Wang, Wang, and Chen}]{hu2022lora}
Edward~J Hu, Yelong Shen, Phillip Wallis, Zeyuan Allen-Zhu, Yuanzhi Li, Shean Wang, Lu~Wang, and Weizhu Chen. 2022.
\newblock \href {https://openreview.net/forum?id=nZeVKeeFYf9} {Lo{RA}: Low-rank adaptation of large language models}.
\newblock In \emph{International Conference on Learning Representations}.

\bibitem[{Hua et~al.(2024)Hua, Li, Luo, Satriadi, Feng, Zhan, Qu, Sharma, Zukerman, Semnani-Azad, and Haffari}]{hua2024sadas}
Yuncheng Hua, Zhuang Li, Linhao Luo, Kadek~Ananta Satriadi, Tao Feng, Haolan Zhan, Lizhen Qu, Suraj Sharma, Ingrid Zukerman, Zhaleh Semnani-Azad, and Gholamreza Haffari. 2024.
\newblock Sadas: A dialogue assistant system towards remediating norm violations in bilingual socio-cultural conversations.
\newblock \emph{arXiv preprint arXiv:2402.01736}.

\bibitem[{Joshi et~al.(2024)Joshi, Dabre, Kanojia, Li, Zhan, Haffari, and Dippold}]{joshi2024natural}
Aditya Joshi, Raj Dabre, Diptesh Kanojia, Zhuang Li, Haolan Zhan, Gholamreza Haffari, and Doris Dippold. 2024.
\newblock Natural language processing for dialects of a language: A survey.
\newblock \emph{arXiv preprint arXiv:2401.05632}.

\bibitem[{Kim et~al.(2022)Kim, Kim, Song, Hwang, and Yeo}]{kim2022botstalk}
Minju Kim, Chaehyeong Kim, Yongho Song, Seung-won Hwang, and Jinyoung Yeo. 2022.
\newblock Botstalk: Machine-sourced framework for automatic curation of large-scale multi-skill dialogue datasets.
\newblock \emph{arXiv preprint arXiv:2210.12687}.

\bibitem[{Li and Liang(2021)}]{li2021prefix}
Xiang~Lisa Li and Percy Liang. 2021.
\newblock Prefix-tuning: Optimizing continuous prompts for generation.
\newblock \emph{arXiv preprint arXiv:2101.00190}.

\bibitem[{Lin(2004)}]{lin2004rouge}
Chin-Yew Lin. 2004.
\newblock Rouge: A package for automatic evaluation of summaries.
\newblock In \emph{Text summarization branches out}, pages 74--81.

\bibitem[{Liu et~al.(2023)Liu, Yang, Jia, Zhang, Zhou, Dai, Yang, and Vosoughi}]{liu2023trainingSociety}
Ruibo Liu, Ruixin Yang, Chenyan Jia, Ge~Zhang, Denny Zhou, Andrew~M Dai, Diyi Yang, and Soroush Vosoughi. 2023.
\newblock Training socially aligned language models in simulated human society.
\newblock \emph{arXiv preprint arXiv:2305.16960}.

\bibitem[{Liu et~al.(2022)Liu, Ji, Fu, Tam, Du, Yang, and Tang}]{ptuning2022}
Xiao Liu, Kaixuan Ji, Yicheng Fu, Weng Tam, Zhengxiao Du, Zhilin Yang, and Jie Tang. 2022.
\newblock \href {https://doi.org/10.18653/v1/2022.acl-short.8} {{P}-tuning: Prompt tuning can be comparable to fine-tuning across scales and tasks}.
\newblock In \emph{Proceedings of the 60th Annual Meeting of the Association for Computational Linguistics (Volume 2: Short Papers)}, pages 61--68, Dublin, Ireland. Association for Computational Linguistics.

\bibitem[{Liu et~al.(2020)Liu, Ott, Goyal, Du, Joshi, Chen, Levy, Lewis, Zettlemoyer, and Stoyanov}]{liu2020roberta}
Yinhan Liu, Myle Ott, Naman Goyal, Jingfei Du, Mandar Joshi, Danqi Chen, Omer Levy, Mike Lewis, Luke Zettlemoyer, and Veselin Stoyanov. 2020.
\newblock \href {https://openreview.net/forum?id=SyxS0T4tvS} {Ro{\{}bert{\}}a: A robustly optimized {\{}bert{\}} pretraining approach}.

\bibitem[{Lytle et~al.(1995)Lytle, Brett, Barsness, Tinsley, and Janssens}]{lytle1995paradigm}
AL~Lytle, JM~Brett, ZI~Barsness, CH~Tinsley, and Maddy Janssens. 1995.
\newblock A paradigm for confirmatory cross-cultural research in organizational-behavior.
\newblock \emph{Research in organizational behavior: an annual series of analytical essays and critical reviews, vol 17, 1995}, 17:167--214.

\bibitem[{Molho et~al.(2020)Molho, Tybur, Van~Lange, and Balliet}]{molho2020direct}
Catherine Molho, Joshua~M Tybur, Paul~AM Van~Lange, and Daniel Balliet. 2020.
\newblock Direct and indirect punishment of norm violations in daily life.
\newblock \emph{Nature communications}, 11(1):3432.

\bibitem[{Morris et~al.(2015)Morris, Hong, Chiu, and Liu}]{morris2015normology}
Michael~W Morris, Ying-yi Hong, Chi-yue Chiu, and Zhi Liu. 2015.
\newblock Normology: Integrating insights about social norms to understand cultural dynamics.
\newblock \emph{Organizational behavior and human decision processes}, 129:1--13.

\bibitem[{Papineni et~al.(2002)Papineni, Roukos, Ward, and Zhu}]{papineni2002bleu}
Kishore Papineni, Salim Roukos, Todd Ward, and Wei-Jing Zhu. 2002.
\newblock Bleu: a method for automatic evaluation of machine translation.
\newblock In \emph{Proceedings of the 40th annual meeting of the Association for Computational Linguistics}, pages 311--318.

\bibitem[{Pfeiffer et~al.(2020)Pfeiffer, Kamath, R{\"u}ckl{\'e}, Cho, and Gurevych}]{pfeiffer2020adapterfusion}
Jonas Pfeiffer, Aishwarya Kamath, Andreas R{\"u}ckl{\'e}, Kyunghyun Cho, and Iryna Gurevych. 2020.
\newblock Adapterfusion: Non-destructive task composition for transfer learning.
\newblock \emph{arXiv preprint arXiv:2005.00247}.

\bibitem[{Pillutla et~al.(2021)Pillutla, Swayamdipta, Zellers, Thickstun, Welleck, Choi, and Harchaoui}]{pillutla2021mauve}
Krishna Pillutla, Swabha Swayamdipta, Rowan Zellers, John Thickstun, Sean Welleck, Yejin Choi, and Zaid Harchaoui. 2021.
\newblock Mauve: Measuring the gap between neural text and human text using divergence frontiers.
\newblock \emph{Advances in Neural Information Processing Systems}, 34:4816--4828.

\bibitem[{Stolcke et~al.(2000)Stolcke, Ries, Coccaro, Shriberg, Bates, Jurafsky, Taylor, Martin, Ess-Dykema, and Meteer}]{stolcke2000dialogue}
Andreas Stolcke, Klaus Ries, Noah Coccaro, Elizabeth Shriberg, Rebecca Bates, Daniel Jurafsky, Paul Taylor, Rachel Martin, Carol~Van Ess-Dykema, and Marie Meteer. 2000.
\newblock Dialogue act modeling for automatic tagging and recognition of conversational speech.
\newblock \emph{Computational linguistics}, 26(3):339--373.

\bibitem[{Van~der Maaten and Hinton(2008)}]{van2008tsne}
Laurens Van~der Maaten and Geoffrey Hinton. 2008.
\newblock Visualizing data using t-sne.
\newblock \emph{Journal of machine learning research}, 9(11).

\bibitem[{Wang et~al.(2019)Wang, Shi, Kim, Oh, Yang, Zhang, and Yu}]{wang2019persuasion}
Xuewei Wang, Weiyan Shi, Richard Kim, Yoojung Oh, Sijia Yang, Jingwen Zhang, and Zhou Yu. 2019.
\newblock Persuasion for good: Towards a personalized persuasive dialogue system for social good.
\newblock \emph{arXiv preprint arXiv:1906.06725}.

\bibitem[{Wen et~al.(2017)Wen, Miao, Blunsom, and Young}]{wen2017latent}
Tsung-Hsien Wen, Yishu Miao, Phil Blunsom, and Steve Young. 2017.
\newblock Latent intention dialogue models.
\newblock In \emph{International Conference on Machine Learning}, pages 3732--3741. PMLR.

\bibitem[{Zhan et~al.(2023{\natexlab{a}})Zhan, Li, Wang, Luo, Feng, Kang, Hua, Qu, Soon, Sharma et~al.}]{zhan2023socialdial}
Haolan Zhan, Zhuang Li, Yufei Wang, Linhao Luo, Tao Feng, Xiaoxi Kang, Yuncheng Hua, Lizhen Qu, Lay-Ki Soon, Suraj Sharma, et~al. 2023{\natexlab{a}}.
\newblock Socialdial: A benchmark for socially-aware dialogue systems.
\newblock \emph{arXiv preprint arXiv:2304.12026}.

\bibitem[{Zhan et~al.(2023{\natexlab{b}})Zhan, Maruf, Qu, Zukerman, and Haffari}]{zhan2023turning}
Haolan Zhan, Sameen Maruf, Lizhen Qu, Ingrid Zukerman, and Gholamreza Haffari. 2023{\natexlab{b}}.
\newblock Turning flowchart into dialog: Plan-based data augmentation for low-resource flowchart-grounded troubleshooting dialogs.
\newblock \emph{arXiv preprint arXiv:2305.01323}.

\bibitem[{Zhan et~al.(2024)Zhan, Wang, Feng, Hua, Sharma, Li, Qu, Azad, Zukerman, and Haffari}]{zhan2024let}
Haolan Zhan, Yufei Wang, Tao Feng, Yuncheng Hua, Suraj Sharma, Zhuang Li, Lizhen Qu, Zhaleh~Semnani Azad, Ingrid Zukerman, and Gholamreza Haffari. 2024.
\newblock Let's negotiate! a survey of negotiation dialogue systems.
\newblock \emph{arXiv preprint arXiv:2402.01097}.

\bibitem[{Zhang et~al.(2019)Zhang, Kishore, Wu, Weinberger, and Artzi}]{zhang2019bertscore}
Tianyi Zhang, Varsha Kishore, Felix Wu, Kilian~Q Weinberger, and Yoav Artzi. 2019.
\newblock Bertscore: Evaluating text generation with bert.
\newblock \emph{arXiv preprint arXiv:1904.09675}.

\bibitem[{Ziems et~al.(2022{\natexlab{a}})Ziems, Yu, Wang, Halevy, and Yang}]{ziems-etal-2022-moral}
Caleb Ziems, Jane Yu, Yi-Chia Wang, Alon Halevy, and Diyi Yang. 2022{\natexlab{a}}.
\newblock \href {https://doi.org/10.18653/v1/2022.acl-long.261} {The moral integrity corpus: A benchmark for ethical dialogue systems}.
\newblock In \emph{Proceedings of the 60th Annual Meeting of the Association for Computational Linguistics (Volume 1: Long Papers)}, pages 3755--3773, Dublin, Ireland. Association for Computational Linguistics.

\bibitem[{Ziems et~al.(2022{\natexlab{b}})Ziems, Yu, Wang, Halevy, and Yang}]{ziems2022moral}
Caleb Ziems, Jane~A Yu, Yi-Chia Wang, Alon Halevy, and Diyi Yang. 2022{\natexlab{b}}.
\newblock The moral integrity corpus: A benchmark for ethical dialogue systems.
\newblock \emph{arXiv preprint arXiv:2204.03021}.

\bibitem[{Ziems et~al.(2022{\natexlab{c}})Ziems, Yu, Wang, Halevy, and Yang}]{ziems2022moralIntegrityCorpus}
Caleb Ziems, Jane~A Yu, Yi-Chia Wang, Alon Halevy, and Diyi Yang. 2022{\natexlab{c}}.
\newblock The moral integrity corpus: A benchmark for ethical dialogue systems.
\newblock \emph{arXiv preprint arXiv:2204.03021}.

\end{thebibliography}
